\documentclass[a4paper,fleqn]{cas-dc}

\usepackage[numbers]{natbib}
\let\printorcid\relax 

\usepackage{graphicx}
\usepackage{caption}
\usepackage{subcaption}
\usepackage{booktabs}
\usepackage{amssymb}
\usepackage{amsmath}
\usepackage{enumitem}
\usepackage{paralist}

\newcommand{\ProjName}{SAIBench}

\begin{document}
\let\WriteBookmarks\relax
\def\floatpagepagefraction{1}
\def\textpagefraction{.001}

\title [mode = title]{SAIBench: A Structural Interpretation of AI for Science Through Benchmarks}                      
\shorttitle{Structural Interpretation of AI for Science}

\shortauthors{Yatao Li et~al.}
 \author[1,2,3]{Yatao Li}
 \cormark[1]
 \ead{yatli@microsoft.com}
 
 \author[1,2]{Jianfeng Zhan}
 \ead{zhanjianfeng@ict.ac.cn}
 
 \address[1]{Institute of Computing Technology Chinese Academy of Science, No.6 Kexueyuan South Road, Haidian Beijing 100190, China}
 \address[2]{University of Chinese Academy of Sciences, No.19(A) Yuquan Road, Shijingshan Beijing 100049, China}
 \address[3]{Microsoft Research, No. 5 Dan Ling Street, Haidian Beijing 100080, China}

\cortext[cor1]{Corresponding author}

\begin{abstract}
Artificial Intelligence for Science (AI4S) is an emerging research field that utilizes machine learning advancements to tackle complex scientific computational issues, aiming to enhance computational efficiency and accuracy. However, the data-driven nature of AI4S lacks the correctness or accuracy assurances of conventional scientific computing, posing challenges when deploying AI4S models in real-world applications. To mitigate these, more comprehensive benchmarking procedures are needed to better understand AI4S models. This paper introduces a novel benchmarking approach, known as structural interpretation, which addresses two key requirements: identifying the trusted operating range in the problem space and tracing errors back to their computational components. This method partitions both the problem and metric spaces, facilitating a structural exploration of these spaces. The practical utility and effectiveness of structural interpretation are illustrated through its application to three distinct AI4S workloads: machine-learning force fields (MLFF), jet tagging, and precipitation nowcasting. The benchmarks effectively model the trusted operating range, trace errors, and reveal novel perspectives for refining the model, training process, and data sampling strategy. 
This work is part of the SAIBench project, an AI4S benchmarking suite.
\end{abstract}


\maketitle

\section{Introduction}


Artificial Intelligence for Science (AI4S) merges machine learning advancements to tackle intricate scientific computational problems, streamlining and accelerating discovery processes. The integration of AI in science enhances computational efficiency and accuracy, solving previously unmanageable problems. This interdisciplinary field can innovate scientific research by developing precise algorithms and models, leading to more accurate scientific conclusions~\cite{argonnenationallaboratoryAIScienceReport}.

AI4S provides many benefits but also presents certain challenges. Traditional scientific computing methods, based on natural laws or approximate theories, enable a deterministic, algorithmic numerical computation process. The precision of these methods is dependent on the theory level used in the calculation. After ensuring accuracy, the focus shifts to correct implementation through rigorous software engineering practices, such as unit and integration tests.

However, AI4S methodologies, characterized by data-driven machine learning techniques, lack the assurance of accuracy found in conventional scientific computing. This lack of guarantees can lead to issues when using AI4S models in real-world scientific applications, potentially causing computation pipeline crashes or misleading scientific conclusions. This unpredictability can deter the widespread use of AI4S methodologies. Thus, it's crucial to establish comprehensive benchmarking procedures to verify AI4S models' reliability, reducing risks and enhancing confidence in using AI in scientific research.
To tackle these challenges, we suggest that the benchmarking method must meet two specific requirements:

\textbf{First, the method must identify the model's trusted operating range within the problem space.} AI4S model correctness cannot be generalized from a few validation samples across the entire problem space. In fact, even if we validate correctness across the entirety of a specific dataset, this doesn't guarantee generalizability. AI4S, designed to aid researchers in exploring the unknown, often faces out-of-distribution problem cases in real-world applications~\cite{liDoesAIScience2023}. Instead of aiming for a difficult or impossible global correctness assertion, we could build a model indicating the AI4S model's correctness confidence level over a specific problem space range.

\textbf{Second, the method must trace errors back to the participating computation components.} The AI4S model's accuracy isn't bounded by scientific theories. Supervised learning provides ground truth signals only at input and output stages, creating a ``black box'' machine learning model, making error tracing challenging. However, numerous components participate in an AI4S workload, including individual sample points in the problem space and various neural network modules. Rather than relying on global error metrics, we should construct an error tracing procedure to isolate issues associated with different AI4S components, like imbalanced data or improper loss function. This way, we can understand how each component impacts accuracy and improve them specifically.

In this paper, we introduce a novel benchmarking approach for AI4S workloads called \textbf{structural interpretation}. This method meets the two key requirements by partitioning both the problem and metric spaces, facilitating a structural exploration of these spaces. The trusted operating range is explored by observing the varying behaviors of a model trained with specific sample points across different problem space regions. Similarly, error tracing is performed by evaluating the model's behavior from various perspectives and analyzing the correlations between input, output, and metrics. We demonstrate this innovative benchmarking method on three AI4S workloads: machine-learning force fields (MLFF), jet tagging, and precipitation nowcasting.
The benchmarks effectively model the trusted operating range and trace errors, revealing new perspectives for model refinement, training process improvement, and data sampling strategy. This showcases the practical use and effectiveness of the proposed structural interpretation approach in assessing the performance and reliability of AI4S workloads.
This work is a component of the \ProjName~project, an AI4S benchmarking suite: \url{https://www.computercouncil.org/SAIBench}.

\vspace{-10px}

\section{Preliminaries of AI for Science Workloads}
The basic workflow for most AI4S workloads is not unlike the traditional machine learning: 
\textbf{1.} Obtain a dataset that consists of input-output pairs: $\mathcal{D}=\{(\hat{x}_i, \hat{y}_i) | \hat{x}_i \in \mathbb{X}, \hat{y}_i \in \mathbb{Y}\}$. Partition the dataset into training, validation and test subsets: $\mathcal{D} = \mathcal{D}_{\mathrm{train}} \cup \mathcal{D}_{\mathrm{val}} \cup \mathcal{D}_{\mathrm{test}}$.
\textbf{2.} Construct a machine learning model: $\mathcal{F}: \mathbb{X} \rightarrow \mathbb{Y}$, where $\mathcal{F}$ can be a neural network or other parameterized models.
\textbf{3.} Design a loss function that describes the difference between the ground truth and prediction: $\mathcal{L}: (\mathbb{Y}, \mathbb{Y}) \rightarrow \mathbb{R}$.
\textbf{4.} Optimize the functional $\mathcal{F}$ over $\mathcal{D}_{\mathrm{train}}$, to minimize $\mathcal{L}$. $\mathcal{D}_{\mathrm{val}}$ does not participate in optimization, but provides a feedback signal to go back to the model construction step.
\textbf{5.} Evaluate the trained model $\mathcal{F}$ over $\mathcal{D}_{\mathrm{test}}$.

The most important difference of AI4S from traditional machine learning workloads is that the problem space is governed by certain natural laws, and therefore ab-initio or approximated theories derived from it. For different AI4S workloads, these theories could participate in different steps in the workflow in various ways, for example, to generate datasets, project the data, or provide physically-informed dataflow in the machine learning model. Different workloads also specify different constraints for the workflow. We now briefly introduce the three workloads we conducted benchmarking on.

\vspace{-10px}
\subsection{Machine-Learning Force Fields}

A force field is a function that calculates forces on each atom in a molecular system, based on atom coordinates and types. The force field plays a central role in molecular dynamics (MD), a vital tool that simulates the atomic motions within chemical systems and enables downstream applications in material science, drug design, computational biochemistry and so on.
Traditionally, the force field is derived from the inter-atomic potential of the molecular system, as the negative gradient of the potential energy: $\vec{f}_i = - \frac{\partial}{\partial \vec{x}_i} E(\vec{x}_1, \vec{x}_2, \dots, \vec{x}_N)$. With the forces computed, the acceleration of each atom can be derived with Newton's Second law: $\vec{f}_i=m_i\vec{a}_i$, and the MD simulation loop incorporates the accelerations to update the velocities of the atoms and displace them repeatedly, forming a trajectory in time-space: $\vec{x'}_i = \vec{x}_i + \vec{v}_i\Delta_t$.

A machine-learning force field (MLFF) offers a data-driven alternative to conventional methods for computing the potential energy and forces. A popular data generation strategy for MLFF is to first compute a trajectory with a fast force field method, and then sample coordinates from the trajectory to label them with an ab-initio force field method governed by Schr\"odinger's equation. The MLFF model then learns from the labeled data with a loss function that combines energy and force errors, for example mean square error (MSE). In the language we previously defined, the workflow is formalized as:
\begin{equation}
  \begin{split}
    \mathcal{D}=\{(\hat{x}_i, \hat{y}_i) | \hat{x}_i = (\mathbf{x}_i, \mathbf{z}_i) , \hat{y}_i = (E_i, \mathbf{F}_i) \} \\
    \mathbf{x}_i, \mathbf{F}_i \in \mathbb{R}^{N \times 3}; E_i \in \mathbb{R}; \mathbf{z}_i \in \mathbb{Z}^N \\
    \mathbf{F}_i = -\frac{\partial}{\partial \mathbf{x_i}} E_i \\
    \widetilde{E}_i, \widetilde{\mathbf{F}}_i = \mathcal{F}(\mathbf{x}_i, \mathbf{z}_i) \\
    \mathcal{L}=\sum_i^{\mathcal{D}_{\mathrm{train}}} [ \alpha (E_i - \widetilde{E_i})^2 + \beta \sum_{j=1}^{N} \sum_{k=x, y, z}(f_{i, j, k}-\widetilde{f}_{i, j, k})^2 ]
  \end{split}
\end{equation}
\subsection{Jet Tagging}

An important scientific task in high energy physics is to study the collision of high energy hadrons. Such experiments are conducted in a particle collider, a large-scale tubular infrastructure that accelerates particle beams confrontationally to near light speed. The collision of two hadrons breaks them into elementary particles like quarks and gluons, which will casecadely collide and form a spray of particles, referred to as a jet.

Different elementary particles initiate different types of jets, and jet tagging identifies the initiating elementary particles from a jet reconstructed from sensory data. Since real-world sensory data is unlabelled, machine-learning based jet taggers rely on a combination of collision simulation, a monte-carlo process governed by quantum chromodynamics (QCD), and accelerator simulation that models the sensor behaviors. The raw sensor reading of a single particle is a 4-momenta tuple that describes the energy and momentum: $(E, p_x, p_y, p_z)$, and it is common to further map the displacement of the particle onto the tubular sensor surface as the pseudorapidity and azimuth angle: $\phi=\mathrm{atan}(p_y/p_x), \eta=\mathrm{atan}(p_z/|\vec{p}|)$, and use the differential representation against the jet vector. The tagging is modeled by a classification process, where the output is an one-hot encoding of the class (e.g. top quark or not).
The machine-learning jet-tagging workflow is then formalized as:
\begin{equation}
  \begin{split}
    \hspace{2cm}
    \mathcal{D}=\{(\hat{x}_i, \hat{y}_i) | \hat{x}_i = (\vec{E}_i, \mathbf{p}_i) , \hat{y}_i = \vec{c}_i \} \\
    \vec{E}_i \in \mathbb{R}^N; \mathbf{p}_i \in \mathbb{R}^{N \times 3}; \sum_j c_{i,j} = 1 \\
    \phi_i=\mathrm{atan}(p_{i,y}/p_{i,x}), \eta=\mathrm{atan}(p_{i, z}/|\vec{p}_i|) \\
    \widetilde{\vec{c}}_i = \mathcal{F}(\vec{E}_i, \vec{\Delta}_{\phi,i}, \vec{\Delta}_{\eta,i}) \\
    \mathcal{L}=\sum_i^{\mathcal{D}_{\mathrm{train}}} \sum_j - c_{i,j}\mathrm{ln}(\widetilde{c}_{i,j})
  \end{split}
\end{equation}

\subsection{Precipitation Nowcasting}

Precipitation nowcasting is a type of weather forecasting that generates quantitative prediction of precipitation over a short range of time, and can provide practical services like just-in-time extreme weather alarming. The input and output for such prediction models are heatmaps that describe the precipitation intensity, organized as time frames. The ground truth is obtained by the radar echo-back field of the clouds. The movement of the atmospheric mass is goverened by the continuity equation, a partial derivative equation that models the interactions between the changes of the quantity in time, space and the residuals: $\frac{\partial \mathbf{x}}{\partial t}+(\mathbf{v}\cdot \nabla)\mathbf{x}=\mathbf{s}$. Different with the previous examples, the governing natural law in this workload is time dependant, and the input and output are consequtive time frames in a weather event, formulated as:
\vspace{-8px}
\begin{equation}
  \begin{split}
    \mathcal{D}=\{(\hat{x}_i, \hat{y}_i) | \hat{x}_i = [\mathbf{x}_{i,-\mathbf{p}} \ldots \mathbf{x}_{i,-1}] , \hat{y}_i = [\mathbf{x}_{i,0} \ldots \mathbf{x}_{i,\mathbf{f}}]\} \\
    \mathbf{x}_{i,j} \in \mathbb{R}^{w \times h} \\
    \frac{\partial \mathbf{x}}{\partial t}+(\mathbf{v}\cdot \nabla)\mathbf{x}=\mathbf{s} \\
    \widetilde{\mathbf{x}}_{i,0} \ldots \widetilde{\mathbf{x}}_{i,\mathbf{f}} = \mathcal{F}(\mathbf{x}_{i,-\mathbf{p}} \ldots \mathbf{x}_{i,-1}) \\
    \mathcal{L}=\sum_i^{\mathcal{D}_{\mathrm{train}}} \sum_{j=0}^{\mathbf{f}} \lVert \widetilde{\mathbf{x}}_{i,j} - \mathbf{x}_{i,j} \rVert 
  \end{split}
  \label{eq:mrms}
\end{equation}
\vspace{-15px}
\subsection{Other AI4S Workloads}
\begin{table*}[t!]
  \begin{center}
    \begin{tabular}{l c c c c c}
    \toprule
       \textbf{Workload}    & \textbf{Input space} & \textbf{Quantitative}  & \textbf{Qualitative}  & \textbf{Time series} \\
       \midrule
       \textbf{MLFF}        & Point cloud          & \checkmark             &                       & \\
       \textbf{Jet Tagging} & Point cloud          &                        & \checkmark            & \\
       \textbf{Precipitation Nowcasting} & Dense matrix & \checkmark        &                       & \checkmark \\
       \midrule
       \textbf{Medical Image Augmentation} & Dense matrix &                 & \checkmark            & \\
       \textbf{Protein Folding} & Point cloud     & \checkmark              &                       & \\
       \textbf{Extreme Weather Detection} & Dense matrix &                  & \checkmark            & \\
       \textbf{Generic PDE Solving} & Dense matrix & \checkmark             &                       & Yes/No \\
       \textbf{Genome Alignment} & Token Sequence & \checkmark             &                        & \\
    \bottomrule
    \end{tabular}
    \end{center}
    \vspace{-10px}
    \caption{Characterization of AI4S Workloads}
    \label{table:workloads}
    \vspace{-15px}
\end{table*}
The characterization of AI4S workloads covers multiple aspects such as input/output formalization, the governing natural law, the prediction task class, etc. Table~\ref{table:workloads} illustrates the characterization of the three benchmarked workloads and a few other representative ones. The ``quantitative'' column indicates numerically meaningful prediction results, like precipitation intensity in $\mathrm{mm}\cdot h^{-1}$ for the precipitation nowcasting task. Conversely, the ``qualitative'' column indicates prediction results serving higher-level semantics, like classification or marking an object's contour. From the table we can see that our selection of three workloads cover typical combinations of characterization perspectives, including popular input space formalizations, both quantitative and qualitative tasks, and time series prediction.

\subsection{Structural Interpretation of AI for Science}
\begin{figure}[t!]\centering
    \includegraphics[width=0.48\textwidth]{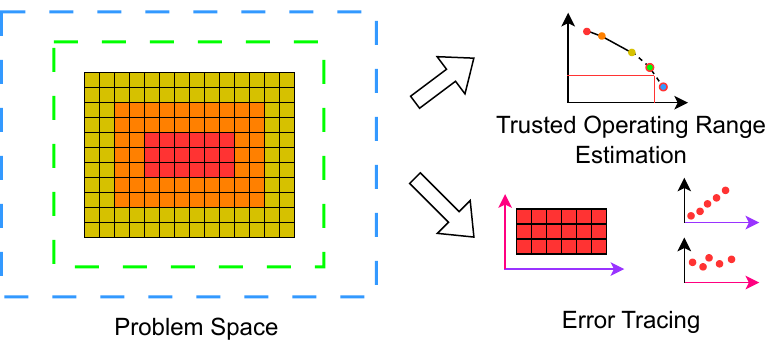}
    \vspace{-20px}
  \caption{Structural Interpretation Methodology}
    \vspace{-20px}
  \label{fig:structural-interpretation}
\end{figure}

We propose structural interpretation to extend the workflow of AI4S, illustrated in figure~\ref{fig:structural-interpretation}. The problem space depicted on the left consists of sampled data points (red, orange and yellow rectangles) in a dataset, and unobserved problem cases beyond it. ``Structural'' means that instead of taking all the data points to train a single model, and aggregate performance metrics, we should construct an evaluation protocol to expose the fine-grained details in both the problem and metric space. The metric space is not limited to the response of a trained model, but can also include features we project directly from the problem space. By having multiple sampling strategies, we obtain a series of evaluations on different slices of the data, and use the correlation between data metrics and model response metrics to extrapolate beyond our full dataset, and construct the trusted operating range. This is demonstrated in the top right plot, where different regions of the problem space are color-coded and tested separately, building up a trend that extrapolates into the green and blue unobserved regions in the true problem space. Similarly, within a single evaluation, we can also break down the aggregated response metrics into the behaviors on individual data points, and study the correlation between the response and the data metrics, and trace the errors in the response metric back to the changes of the data metrics. This is depicted in the bottom right of the figure, where the ``red'' data points exhibit two different feature metrics, and both are analyzed against the same error metric, showing that the error of the model is more sensitive to one metric.

\vspace{-10px}
\section{Related works}

Numerous benchmarking suites have been developed to evaluate machine learning workloads~\cite{gaoAIBenchScalableComprehensive2019,gaoAIBenchScenarioScenarioDistilling2021,mattsonMLPerfTrainingBenchmark}. 
Machine learning model benchmarks emphasize convergence rate and accuracy and focus on metrics such as the validation and test loss function performance, while industrial AI benchmarks prioritize cost-efficient model deployment~\cite{gaoAIBenchIndustryStandard2019} and concentrate on performance metrics such as model throughput and hardware resource utilization. 
Recently, benchmarks specifically targeting AI for science are also proposed~\cite{thiyagalingamScientificMachineLearning2022}, which carries over the methodologies of conventional AI benchmarking.
AI4S research fields also propose to evaluate the machine learning models with benchmarks, which are usually datasets computed by conventional scientific computing methods\cite{chmielaMachineLearningAccurate2017,christensenRoleGradientsMachine2020,kasieczkaMachineLearningLandscape2019,unkePhysNetNeuralNetwork2019} or collected from real-world sensor readings~\cite{zhangMultiRadarMultiSensorMRMS2016, racahExtremeWeatherLargescaleClimate2017}.
Another emerging approach to better understand the behavior of AI is the explainable AI(xAI)~\cite{gohelExplainableAICurrent2021}, which inserts probes into hidden layers of the machine learning model, and investigate the relationship between the probe signals and the semantic of the input/output. The success of this approach depends on clear definition of semantic, which is relatively straightforward to obtain in case studies in conventional AI workloads such as natural language understanding. However, in scientific computing, the numerical values reflecting the governing natural law are often beyond human comprehension for both conventional and machine-learning methods.

\vspace{-10px}
\section{Benchmarking MLFF}

The benchmark focuses on evaluating NequIP \cite{batznerEquivariantGraphNeural2022}, an equivariant graph neural network architecture specifically designed for learning MLFF from ab-initio calculations. 
NequIP emerges as a prominent representative of the state-of-the-art in this field, distinguished by its exceptional data efficiency and superior performance when compared to previous HDNN-style neural networks \cite{behlerGeneralizedNeuralNetworkRepresentation2007} and kernel-based methods \cite{bartokGaussianApproximationPotentials2010}. NequIP's remarkable data efficiency enables accurate modeling of MLFF with minimal training data, making it an appealing choice for scenarios where data availability is limited or costly to obtain.

To assess the performance of this model, we utilize the revised MD17 (rMD17) dataset \cite{christensenRoleGradientsMachine2020}, which originated from the MD17 dataset \cite{chmielaMachineLearningAccurate2017} that comprises data points obtained from MD simulation trajectories based on density functional theory (DFT), encompassing a predefined set of molecules. The rMD17 dataset further enhances the MD17 dataset by employing a more accurate level of theory, thereby mitigating numerical noise and improving data quality.
It is important to call out that we adopt a different setup compared to conventional machine-learning-style practices. In a conventional machine-learning-style setup, it would assume that the data points (from both the MD17 and rMD17 datasets) are randomly sampled from the ground truth problem space. Training and testing are subsequently conducted by randomly partitioning the data into subsets. 
However, MD17 data points are drawn from simulated trajectories, resulting in inherent correlations in the time domain. Consequently, randomly sampling training and test subsets can lead to the interleaving of data points from different time steps. While this scenario aligns with the ideal situation in MLFF-powered MD, where simulated data covers a wide range of molecule conformation space, we argue that it is not the case for the MD17/rMD17 dataset, which will be demonstrated in the forthcoming experimental results.
The rMD17 dataset is often consumed in a random train/test split manner, as mentioned before because the data points are not ordered as a trajectory time series. This can be mitigated by sorting the data with the ``old\_index'' field, which maps the data points back to the original MD17 and restores temporal order in the data. Our benchmarking fixture is established on this calibrated dataset by splitting out the last 10\% data in the time series as the test subset.

\vspace{-8px}
\subsection{Sample Efficiency}

\begin{figure}\centering
   \begin{subfigure}{0.23\textwidth}
     \centering
     \includegraphics[width=\linewidth]{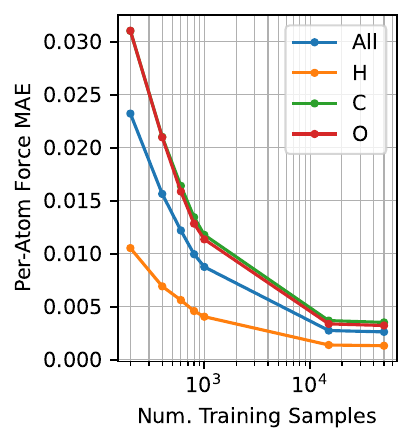}
     \caption{Sample Efficiency (Forces)}\label{fig:mlff-1a}
   \end{subfigure}
   \hfill
   \begin{subfigure}{0.24\textwidth}
     \centering
     \includegraphics[width=\linewidth]{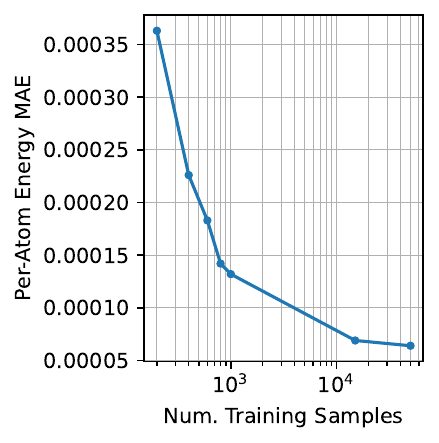}
     \caption{Sample Efficiency (Energy)}\label{fig:mlff-1b}
   \end{subfigure}
   \caption{Sample Efficiency Benchmarks}
    \vspace{-15px}
   \label{fig:mlff-1}
\end{figure}

We evaluate the sample efficiency of the model by fixing the training data window to the first 90\% of the trajectory simulated on an aspirin molecule and progressively sample more data (200, 400, 600, 800, 1000, 15000, and 50000 samples, respectively) from the window into different training subsets, and compare the performance of trained models on the test subset. The training process for each subset is given a fixed wall time budget, allowing all to converge properly. We compare both per-atom force mean average error (MAE) and per-atom energy MAE for the trained models. 
The benchmarking results are illustrated in figure~\ref{fig:mlff-1}.

We can see that the model has good sample efficiency, achieving per-atom energy MAE of less than 4 meV over the test data given only 200 training samples. More specifically, given a fixed training data window, the performance of the trained models progressively improves with more training data points. Both the performance on energy and forces follow a similar trend, where increasing the number of samples results in proportional improvements up to 1000 samples, but the gain decreases exponentially afterward, where the benefit of increasing the size of the training set from 1K to 15K samples is not as good as increasing from 200 to 400, albeit at the cost of much longer computational cost spent in each training epoch. 

In conventional AI benchmarking, it is not practical to evaluate the performance on parts of each data point, such as the first ten tokens generated from a model or the accuracy of prediction about the top-left part of an image. However, the structural and composable nature of molecular data allows for a more versatile projection of performance results, providing a unique opportunity to evaluate AI performance in multiple dimensions and giving more insights into the model's behavior and capabilities.
For example, Figure~\ref{fig:mlff-1a} additionally presents per-atom forces MAE for each species of atoms (Hydrogen, Carbon, and Oxygen). The analysis reveals that the error on different species generally follows the same trend, and the error on Hydrogen is significantly lower than the others due to its low atomic charge. Interestingly, the errors are not strictly proportional to the atomic charge of each species, as one might expect the errors of Oxygen to be proportionally higher than that of Carbon, but the data shows otherwise.

This observation suggests that force prediction is sensitive to the structural configuration of the molecule in addition to the invariant features of each atom. Moreover, it indicates that the model captures more structural information to reflect the steepness in the potential energy surface than an empirical potential energy equation. 

\vspace{-8px}
\subsection{Time-series Extrapolation}
The previous benchmark evaluates the model performance when the entire range of trajectory up to the test window is available to the training process. That is, the model is trained on data sampled from 9 times more time steps (90\%) to predict the immediately upcoming steps (10\%). 
In real-world MLFF-powered MD simulations, it is expected that the MLFF model should be able to support longer runs with more time steps, where the training window might not cover a large number of time steps compared to the inference steps and may not be immediately adjacent to the inference window.
To evaluate the model's performance under such conditions, multiple variations of benchmarks are created using a grid-scan method to vary the size of the training window and its starting point. The training window sizes are set to 30\%, 45\%, 60\%, 75\%, and 90\% of the whole trajectory, while the starting points are set to 0\%, 15\%, 30\%, 45\%, and 60\% of the whole trajectory. For each of these training window variants, we train a model with 15K data points sampled from the window, respectively, and their performance is tested on the final 10\% of the trajectory.

\begin{figure}[t!]\centering
    \includegraphics[width=0.48\textwidth]{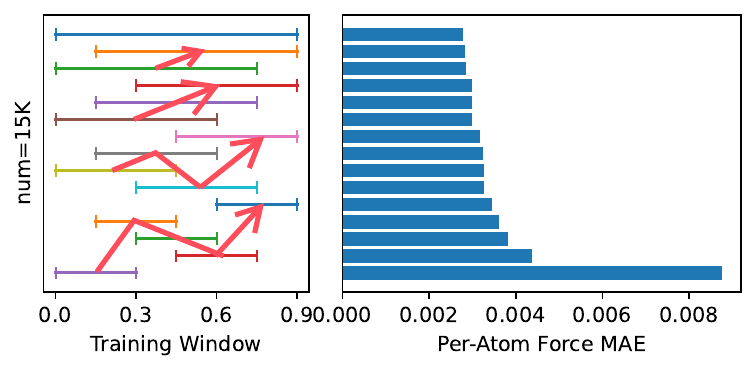}
    \vspace{-20px}
  \caption{Time-Series Extrapolation Benchmarks}
    \vspace{-20px}
  \label{fig:mlff-2.1}
\end{figure}

Figure~\ref{fig:mlff-2.1} presents the time-series extrapolation benchmarking results. Each horizontal line segment in the left part of the chart represents a training window variant, with starting/ending points in the 0\%-90\% range. The bar on the right corresponds to the per-atom force MAE evaluated on the test window for the model trained on this specific training window. The data is sorted by test performance, with the training window on the first row having the best test performance.
The data shows that the test performance varies significantly with different training windows, and the best window is the widest (0\% to 90\%) and the worst is the smallest window temporally distant from the test window (0\%-30\%). This observation suggests that maintaining the model's accuracy over a long trajectory is challenging, as it may not have enough information from distant data to generalize effectively to the test window.

\begin{figure}\centering
  \includegraphics[width=0.275\textwidth]{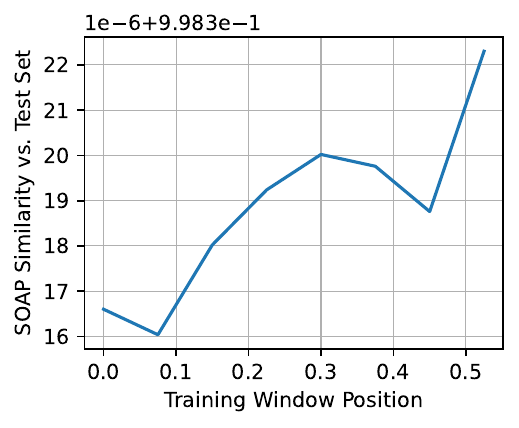}
    \vspace{-10px}
  \caption{Training window vs. Test window SOAP similarity}
    \vspace{-20px}
  \label{fig:mlff-soap}
\end{figure}

Furthermore, it is observed that the model responses exhibit a pattern where, given a fixed window starting position (same starting point on X axis), the performance increases monotonically with the window size. However, given a fixed window size (same segment length in the left part), the performance does not monotonically increase as the training window moves closer to the testing window.
Instead, the performance follows a pattern that longer windows are always better than shorter ones, and the windows with the same length follow a zig-zag trend marked with red arrows.
To understand why this occurs, the SOAP (Smooth Overlap of Atomic Positions) descriptor \cite{bartokRepresentingChemicalEnvironments2013} is leveraged. The SOAP descriptor computes a high-dimensional feature vector for a given molecular system, allowing for the comparison of different molecular configurations. The correlation between two molecular configurations can be calculated by computing the cosine similarity of their corresponding SOAP descriptors.
By computing all pairwise correlations between the training windows with a 30\% range and the testing window, the mean average values are used to represent the similarity of the training windows to the test window. This information is visualized in figure~\ref{fig:mlff-soap}.

The curve depicted in the figure shows that similarity does not monotonically increase as the training window gets closer to the testing window, offering two key insights.
Firstly, it indicates that the trajectory does not perpetually drift from the initial molecular configuration, but occasionally reverts to earlier trajectories' data distribution. This behavior suggests the MD simulation may display periodicity or recurring molecular configuration patterns, crucial for understanding the system's underlying dynamics.
Secondly, the result reveals a clear relationship between the window similarity metric and test performance, suggesting a potential accuracy indicator in a real-world MLFF-powered MD system. A similarity drop below a certain threshold could signal that the MLFF-powered MD loop is heading towards out-of-distribution space, potentially necessitating further model fine-tuning to maintain accuracy and stability.

\vspace{-8px}
\subsection{Error Tracing}

\begin{figure}\centering
   \begin{subfigure}{0.235\textwidth}
     \centering
     \includegraphics[width=\linewidth]{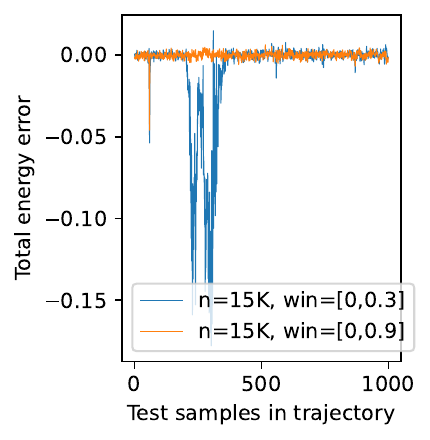}
    \vspace{-20px}
     \caption{Different Windows}\label{fig:mlff-3a}
   \end{subfigure}
   \begin{subfigure}{0.235\textwidth}
     \centering
     \includegraphics[width=\linewidth]{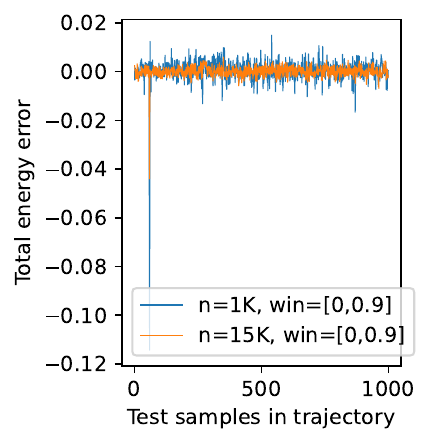}
    \vspace{-20px}
     \caption{Different Sizes}\label{fig:mlff-3b}
   \end{subfigure}
    \vspace{-5px}
  \caption{Performance Impact of Different Sampling Strategies}
    \vspace{-20px}
  \label{fig:mlff-timeline}
\end{figure}

We examine the error structure by breaking down the statistical results and projecting performance metrics on individual test samples from different perspectives.
The iterative nature of the molecular dynamics trajectory allows us to examine error development over time. Additionally, we compare the performance impact induced by different sampling strategies for constructing training subsets. We first fix the training set size to 15K, studying the total energy error for all 1000 test data points for different sampling windows. We then fix the sampling window to the full training set (first 90\% of the trajectory) and study the total energy error for different training set sizes (1K vs. 15K). The result is shown in figure~\ref{fig:mlff-timeline}. Both subfigures show that the total energy error for most test samples centers around 0, indicating unbiased predictions.
However, figure~\ref{fig:mlff-3a} reveals that for the narrow sampling window (blue), a time range exists during which the predictions are consistently negatively biased, contributing significantly to the mean average energy error. The close time correlation indicates that this error type is sensitive to the molecular system's geometric configuration, with the narrow sampling window failing to cover similar configurations, leading to significant errors.
Another error type that persists across both\ref{fig:mlff-3a} and\ref{fig:mlff-3b} is the negative spike at the test time series start. Increasing the sampling window size does not eliminate that spike, but increasing sampling strength from 1K to 15K significantly lowers the spike, consistent with the error reduction on all other test samples.
The implication is that the sampling window and strength lead to two independent error sources: the geometric space-sensitive error and the stiff potential surface-induced error. The former introduces severe errors for a narrow range in the time series, and the latter affects the global ``amplitude'' of error for all samples.

\begin{figure}\centering
   \begin{subfigure}{0.22\textwidth}
     \centering
     \includegraphics[width=\linewidth]{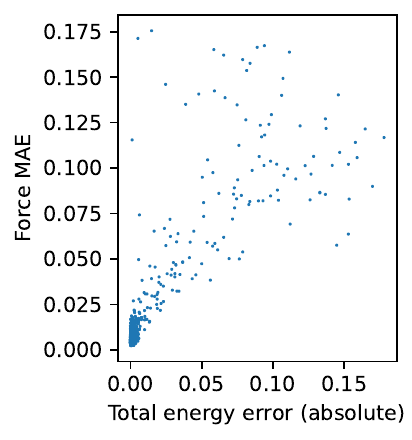}
    \vspace{-20px}
     \caption{1K Samples}\label{fig:mlff-4a}
   \end{subfigure}
   \begin{subfigure}{0.22\textwidth}
     \centering
     \includegraphics[width=\linewidth]{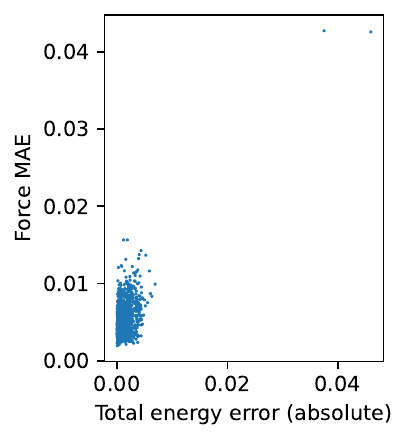}
    \vspace{-20px}
     \caption{15K Samples}\label{fig:mlff-4b}
   \end{subfigure}
    \vspace{-5px}
  \caption{Correlation Between Energy and Force Error}
    \vspace{-15px}
  \label{fig:mlff-energy-force}
\end{figure}

We are interested in the prediction error correlation on the energy and forces. Figure~\ref{fig:mlff-energy-force} shows the scatter plots of energy vs. force error of two different models, trained with 1K and 15K data points sampled from the full training trajectory. The aim of selecting these two models is to compare behavior with and without geometric-sensitive errors. From both subfigures, we observe a pattern in the energy vs. force error distribution: 1) a main body where both errors are low and show little correlation, and 2) for higher errors, a linear correlation between force and energy error. This is attributable to the MLFF model training's algorithmic workflow, where the loss function comprises the energy and force errors, and the force is computed as the energy gradient. This effectively results in the model's weights being used twice to optimize against two different objectives, resulting in a compromise balancing the two objectives. When the error is low, the molecule configuration falls into the well-optimized high-dimensional space of the hidden representation. However, when the energy error is high, most neurons are inactive and cannot pick up the input features, leading to the weights being reused linearly.

\vspace{-15px}
\section{Benchmarking ML-based Jet Tagging}

The benchmark of ML-based jet tagging uses ParticleTransformer~\cite{quParticleTransformerJet2022}, a self-attention permutation invariant neural network architecture designed for learning a high energy particle classifier. Unlike conventional transformer architectures, it omits positional encoding for each input token to achieve permutation invariance, and introduces physical prior through a pairwise particle interaction matrix. With the removal of positional encoding, all embedded particle inputs undergo position-agnostic query and key space mappings, with the self-attention process resembling message gathering in a fully-connected graph convolutional network. This approach, compared to previous ones like point cloud-based ParticleNet~\cite{quParticlenetJetTagging2020}, expands the perception field by not restricting connectivity with a top-k locality graph, yielding better performance.
To assess the performance of this model, we use the TopLandscape~\cite{kasieczkaMachineLearningLandscape2019} dataset, which contains 1 million signal and 1 million background jets. We follow the standard protocol of the dataset to first split the full dataset into training, validation and test sets that consist of 600k, 200k and 200k of signal and background jets respectively. We then conduct structural interpretation of the model by using different sampling strategies from the training and test sets.

\vspace{-8px}
\subsection{Sample Efficiency}

\begin{figure}\centering
   \begin{subfigure}{0.225\textwidth}
     \centering
     \includegraphics[width=\linewidth]{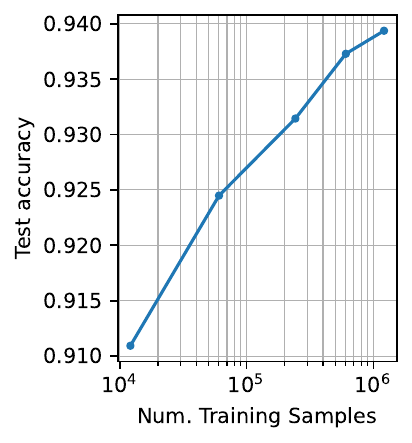}
    \vspace{-20px}
     \caption{Accuracy}\label{fig:jet-1a}
   \end{subfigure}
   \begin{subfigure}{0.225\textwidth}
     \centering
     \includegraphics[width=\linewidth]{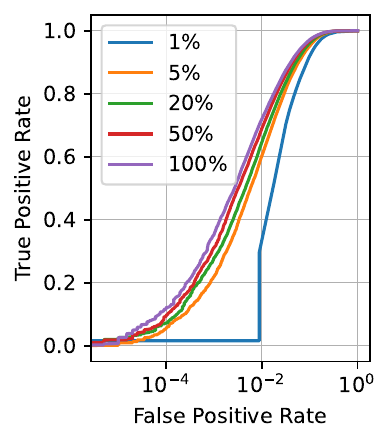}
    \vspace{-20px}
     \caption{ROC}\label{fig:jet-1b}
   \end{subfigure}
    \vspace{-10px}
  \caption{Jet-Tagging Sample Efficiency}
    \vspace{-20px}
  \label{fig:jet-fig1}
\end{figure}

We first examine the model's sample efficiency. Unlike the time-correlated molecular dynamics trajectories, a jet tagging dataset comprises independent events. We randomly sample 1\%, 5\%, 20\%, 50\%, and 100\% of the training samples, train models on these subsets for a constant number of batches, and then test these models on the full test set, as shown in figure~\ref{fig:jet-fig1}. \ref{fig:jet-1a} presents the test set classification accuracy of the models. Contrary to the molecular dynamics sample efficiency test, the jet tagging task displays a consistent log-linear trend, with no accuracy gain drop at higher sampling probabilities, suggesting that utilizing a larger dataset can further enhance performance.
The top quark/background jet tagging is a binary classification task, allowing us to apply a ``gain'' on the classification confidence. A high gain results in the model detecting more signals but with more false positives, while a low gain reduces noise at the cost of fewer true positives. The model's continuous behavior change can be plotted as an ROC curve, as shown in~\ref{fig:jet-1b}. While the test accuracy follows a log-linear trend and performance seems reasonable for the model trained with the 1\% subset, it displays a sharp discrepancy in the ROC curve, indicating that the model cannot suppress the false positive rate below $10^{-2}$, thus exhibiting a high background noise level.

\vspace{-8px}
\subsection{Physical Prior Injection Test}
The particles produced in a collision are detected by the sensors surrounding the collider, forming a cloud of points.
The raw representation of each point is a 4-momenta vector $(E, p_x, p_y, p_z)$.
Conventionally, the neural networks for jet tagging tasks use a physically informed pre-processing step that maps the raw 4-vectors to a tubular surface, represented as 2D vectors $(d_{\phi}, d_{\eta})$.
It is desirable to study if such physical prior injection is always beneficial to the model performance, or the raw data is better if the training dataset is sufficiently large.

\begin{figure}\centering
   \begin{subfigure}{0.228\textwidth}
     \centering
     \includegraphics[width=\linewidth]{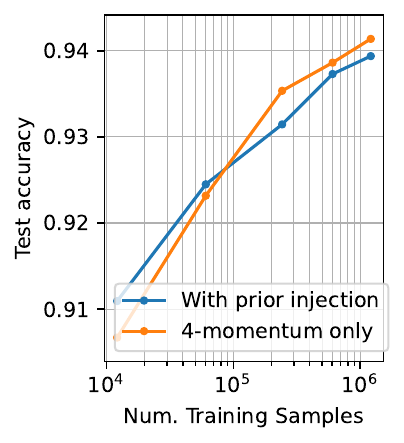}
    \vspace{-20px}
     \caption{Accuracy}\label{fig:jet-3a}
   \end{subfigure}
   \begin{subfigure}{0.228\textwidth}
     \centering
     \includegraphics[width=\linewidth]{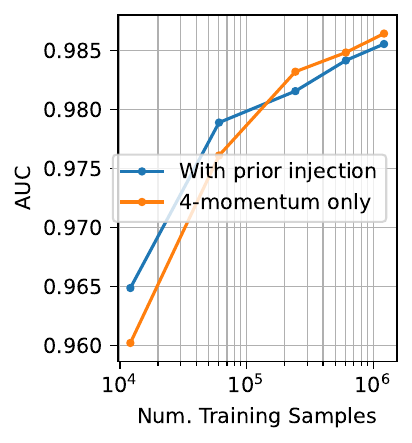}
    \vspace{-20px}
     \caption{AUC}\label{fig:jet-3b}
   \end{subfigure}
    \vspace{-5px}
  \caption{Effectiveness of Physical Prior Injection}
    \vspace{-20px}
  \label{fig:jet-physical}
\end{figure}

Figure~\ref{fig:jet-physical} shows the performance of models trained with the physical prior injections and ones with the raw 4-momenta vectors as the input features. We present the accuracy of the models in~\ref{fig:jet-3a}, and fold the ROC curves into AUC metrics, as shown in~\ref{fig:jet-3b}.
From the figure above we can see that physical prior injection indeed outperformed the raw data with less data.
But as the training data set size grows, the raw data took over, suggesting that the model developed its internal representation that is superior than the straightforward prior injection.

\begin{figure*}\centering
  \includegraphics[width=0.98\textwidth]{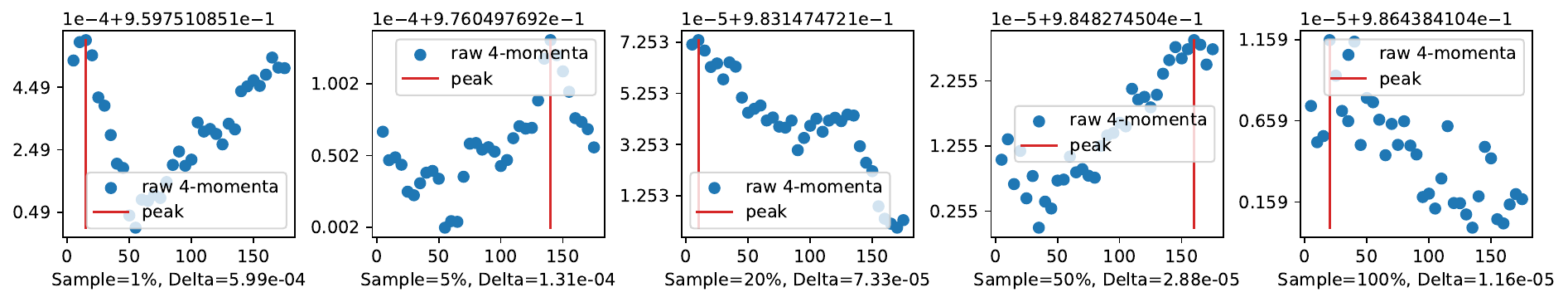}
    \vspace{-11px}
  \caption{Rotation Invariance Test}
    \vspace{-24px}
  \label{fig:jet-rotation}
\end{figure*}

\vspace{-8px}
\subsection{Rotational Invariance Test}
In the previous experiment, we evaluated models trained with raw 4-momenta vectors and physically informed inputs. While physically informed inputs are rotationally invariant, raw 4-vectors are not, leading us to investigate the performance impact of rotation and whether the raw 4-vectors' performance gains can offset the errors in rotated data.
Notably, while it's possible to apply a Lorentz transformation to the dataset, the actual experiment data is obtained from the detector, which should be considered a zero-momentum reference frame. Therefore, we use conventional Newtonian rotational transformations. To match the physical simulation setup, we limit the rotation axis to the direction the hadrons are accelerated, creating 36 rotated test set versions, incrementing by 5 degrees each and covering 180 degrees in total.
Figure\ref{fig:jet-rotation} shows the AUC performance of the models trained with different data amounts, tested against rotation-transformed versions of the test sets. All models exhibit differing performance responses to rotated test sets, establishing a periodic pattern. Also, a model's performance on a specific rotated test set depends on the alignment of the rotated test set and the sampled training set - it's not always the case that models perform best on the non-rotated test set. Cross-checking the performance readings with figure\ref{fig:jet-3b} reveals that the variance caused by the rotational data is well compensated by the performance gain of using the raw 4-momenta vectors because even the worst case for raw 4-momenta models outperforms the ``baseline'' models trained with physical injected prior for larger training subsets significantly. We also report the performance difference between the best and worst rotational dataset in each plot, shown in the X-axis label, revealing that the delta value steadily decreases with increasing training set size. This suggests that using raw 4-momenta does indeed cause rotational variance, but the variance can be mitigated by adding more training data, which is a prerequisite for outperforming engineered physical prior features.

\vspace{-8px}
\subsection{Out-of-Distribution Jet Energy Test}
\begin{figure}\centering
   \begin{subfigure}{0.23\textwidth}
     \centering
     \includegraphics[width=\linewidth]{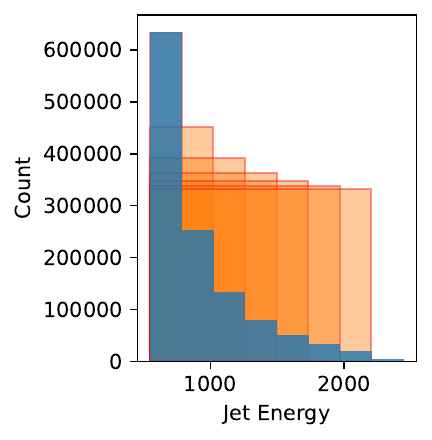}
    \vspace{-20px}
     \caption{Train}\label{fig:jet-5a}
   \end{subfigure}
   \begin{subfigure}{0.23\textwidth}
     \centering
     \includegraphics[width=\linewidth]{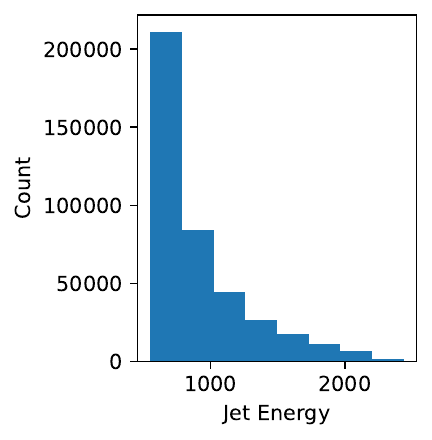}
    \vspace{-20px}
     \caption{Test}\label{fig:jet-5b}
   \end{subfigure}
  \caption{Jet Energy Distribution}
    \vspace{-10px}
  \label{fig:jet-energy-dist}
\end{figure}

An important use case for the jet tagging classifier is to detect rare events during particle collisions. Thus, it is essential to examine a jet tagging machine learning model's out-of-distribution performance to see if it can detect rare events deviating significantly from the training set. The TopLandscape dataset comprises events with a total jet energy ranging from 550 to 2440 GeV, and we partition both the training and test sets by the total jet energy into 8 bins, each with an equal width of 236.25 GeVs. The out-of-distribution evaluation for jet energy involves training a model on lower energy bins and testing performance on the rest. Figure~\ref{fig:jet-energy-dist} displays the jet energy distribution for both training and test sets.
Both training and testing data follow a long tail distribution of jet energy, with the first bin containing approximately half of the samples.
Our data efficiency experiments show the model performance grows log-linearly with the training dataset size. This differs from molecular dynamics experiments, where the model performance grows only linearly with the training dataset size after reaching a certain data volume.
This means we cannot simply combine different bins together because datasets with different sizes will significantly affect the model accuracy. Instead, we must sample an equal number of samples from combined bins. The sampling ranges and probabilities for the training sets are illustrated by the orange rectangles in~\ref{fig:jet-5b}, with each training set covering a wider range of jet energies, and lower sampling probabilities uniformly across the sampled bins.

\begin{figure}\centering
   \hspace{8px}
   \begin{subfigure}{0.38\textwidth}
     \centering
     \includegraphics[width=\linewidth]{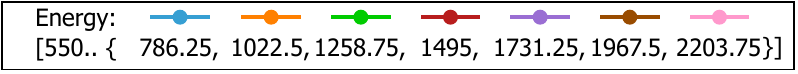}
   \end{subfigure}

   \hspace{-10px}
   \begin{subfigure}{0.21\textwidth}
     \centering
     \includegraphics[width=\linewidth]{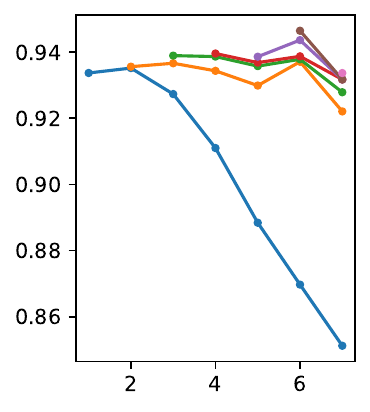}
    \vspace{-20px}
     \caption{Accuracy}\label{fig:jet-6a}
   \end{subfigure}
   \begin{subfigure}{0.21\textwidth}
     \centering
     \includegraphics[width=\linewidth]{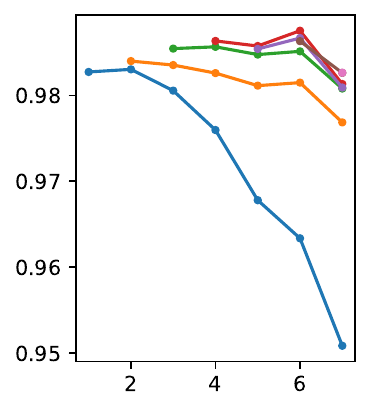}
    \vspace{-20px}
     \caption{AUC}\label{fig:jet-6b}
   \end{subfigure}
  \caption{Out-of-Distribution Jet Energy Test}
    \vspace{-10px}
  \label{fig:jet-energy-ood}
\end{figure}

For each model trained with a specific jet energy range, we iteratively test its performance on the test set bins with out-of-distribution jet energy. The accuracy and AUC performance of the models are shown in figure~\ref{fig:jet-energy-ood}. Both accuracy and AUC metrics display a quadratic decaying trend as the test bin deviates from the training bins. An exception is bin 6 (1967.5-2203.75), where a notable performance bump is observed for all models. This suggests that we can construct a safe operating range model for the trained model based on energy deviation, similar to using the SOAP similarity for the MLFF model.

\vspace{-8px}
\subsection{Error Tracing}
\begin{figure}\centering
   \begin{subfigure}{0.23\textwidth}
     \centering
     \includegraphics[width=\linewidth]{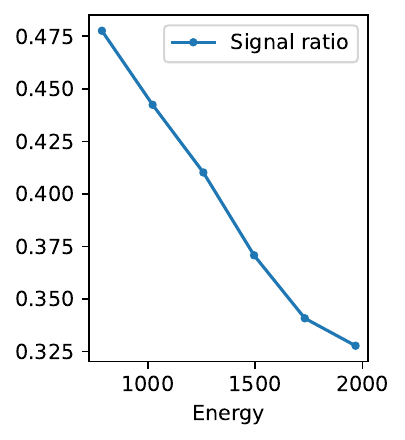}
    \vspace{-20px}
     \caption{Signal Ratio in Test Bins}\label{fig:jet-7a}
   \end{subfigure}
   \begin{subfigure}{0.23\textwidth}
     \centering
     \includegraphics[width=\linewidth]{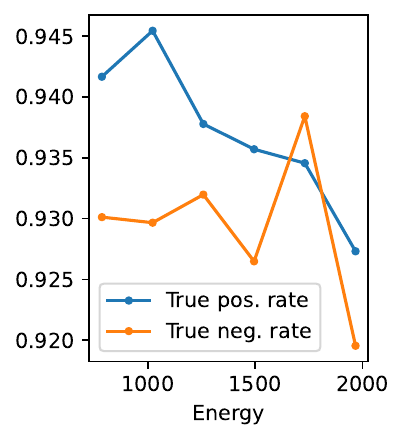}
    \vspace{-20px}
     \caption{Accuracy Breakdown}\label{fig:jet-7b}
   \end{subfigure}
  \caption{Signal and Background Accuracy Breakdown}
    \vspace{-20px}
  \label{fig:jet-error-breakdown}
\end{figure}
To better comprehend the ``bump'' in the accuracy and AUC response, we examine the performance in correctly identifying the signal and background respectively. Figure~\ref{fig:jet-error-breakdown} illustrates the break-down accuracy on signal vs. background for the model trained on samples from the two lowest energy bins. Figure~\ref{fig:jet-7a} displays the signal ratio in the test bins, showing a linear decrease as energy increases. This implies that the background accuracy weight becomes higher when calculating overall accuracy. Figure~\ref{fig:jet-7b} indicates that while the signal accuracy response aligns with our previous observations, it is the background accuracy that gets boosted in bin 6. There is also a gap between signal and background accuracy, suggesting that in real-world deployment where the signal-background ratio differs from the standard protocol in the TopLandscape dataset, performance metrics will vary.

\vspace{-10px}
\section{Benchmarking Precipitation Nowcasting}

The benchmark of precipitation nowcasting uses NowcastNet~\cite{zhangSkilfulNowcastingExtreme2023}, which combines a physically-informed neural network encoder approximating the continuity equation, with a GAN-based decoder augmenting the physics-based result with rich prediction details. Adhering to physics constraints, atmospheric activity prediction is decomposed into predicting the atmospheric mass's velocity field and the residual component representing precipitation intensity. Further augmented with a GAN-based decoder, the model generates physically plausible, detail-rich predictions referred to as "skilful nowcasting".
We assess this neural network architecture using the trained model weights and the test subset of the multi radar multi sensor (MRMS) dataset~\cite{zhangMultiRadarMultiSensorMRMS2016} published by~\cite{zhangSkilfulNowcastingExtreme2023}. The test set includes 100000 precipitation events, each with 29 consecutive 10-minute interval time frames. Each frame is a 512x512 grid representing the precipitation intensity over the US detected by radar echo-back wave. The model takes the first 9 frames as input, generating 20 frames representing fine-grained precipitation intensity predictions for over 3 hours.

\begin{figure}\centering
  \includegraphics[width=0.48\textwidth]{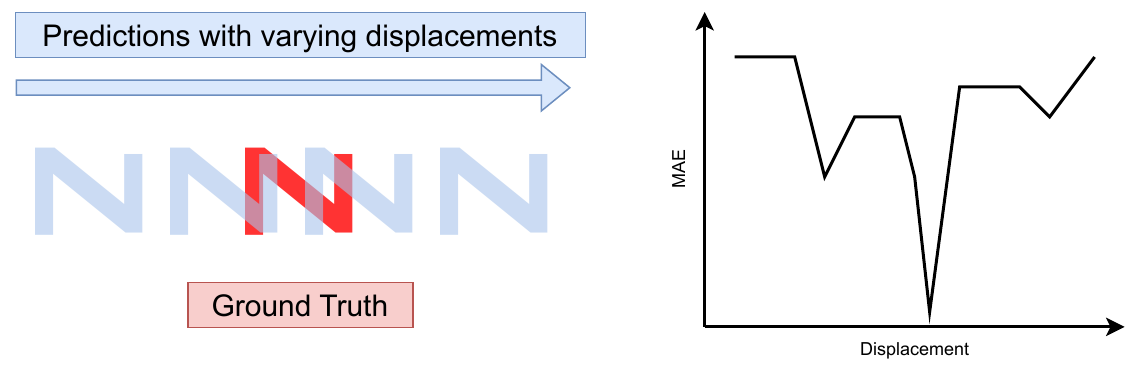}
    \vspace{-15px}
  \caption{MAE of Predictions With Varying Displacements}
    \vspace{-15px}
  \label{fig:mrms-moving-mae}
\end{figure}

The structural interpretation of an AI4S workload requires evaluation metrics that are statistically feasible and scientifically meaningful. However, such metrics are lacking in AI-powered precipitation analysis. Domain-specific metrics like critical success index study model behaviors on individual events. Previous works~\cite{ravuriSkilfulPrecipitationNowcasting2021,zhangSkilfulNowcastingExtreme2023} only present CSI scores for a limited number of events, without generalizing this metric over the entire dataset. Traditional machine learning metrics such as MAE and PSNR are used in motion image prediction~\cite{wangPredRNNRecurrentNeural2017}, but they don't accurately represent precipitation prediction quality. These metrics don't consider atmospheric physics, focusing instead on pixel-to-pixel error, incorrectly attributing prediction errors. For example, a model may correctly predict the precipitation-causing residual but incorrectly predict the atmospheric mass's velocity field, resulting in a same-shaped, displaced intensity map. Conventional metrics can't identify such alignment issues and would assign non-smooth scores for partially-aligned predictions or be insensitive to totally misaligned cases. These issues are illustrated in figure~\ref{fig:mrms-moving-mae}. Therefore, our immediate task is to design scientifically meaningful evaluation metrics for machine learning-based precipitation analysis.

\vspace{-8px}
\subsection{Cumulative Critical Success Index}
The critical suceess index (CSI) is an evaluation metric widely used in precipitation analysis~\cite{hoganDeterministicForecastsBinary2011} and adapted by state-of-the-art machine learning-based precipitation analysis solutions. The CSI metric requires a parameter $T$ to binarilize the precipitation intensity into events and non-events. The threshold is then applied to both the ground truth and prediction of all local areas, where:
\begin{equation}
A=\sum_{\substack{\mathrm{gt}>=T \\ \mathrm{pd}>=T}}{1}, B=\sum_{\substack{\mathrm{gt}>=T \\ \mathrm{pd}<T}}{1}, C=\sum_{\substack{\mathrm{gt}<T \\ \mathrm{pd}>=T}}{1},
\mathrm{CSI}=\frac{A}{A+B+C}
\end{equation}
 The choice of threshold value $T$ is crucial in characterizing the predictions, and some previous works have used multiple thresholds to describe the prediction capabilities of light, medium and heavy precipitation, for example at $16\mathrm{mm}\cdot h^{-1}$, $32\mathrm{mm}\cdot h^{-1}$, $64\mathrm{mm}\cdot h^{-1}$ respectively, and compute the average performance: $\mathrm{CSI}_{\mathrm{avg}}=(\mathrm{CSI}_{16}+\mathrm{CSI}_{32}+\mathrm{CSI}_{64})/3$. Simply put, this metric computes the ratio of correctly predicted events (true positive) to all predictions except the correctly predicted non-events (true negative) over the global area.
\begin{figure}\centering
  \includegraphics[width=0.48\textwidth]{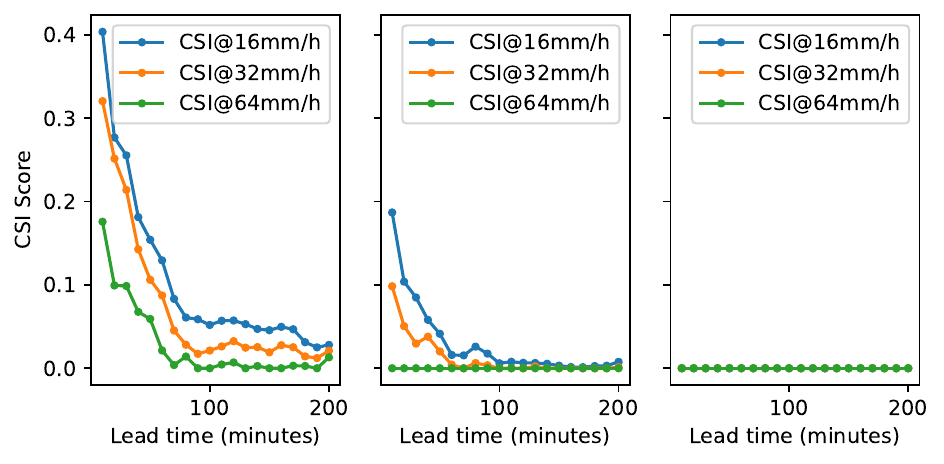}
    \vspace{-15px}
  \caption{CSI Scores of Different Precipitation Events}
    \vspace{-15px}
  \label{fig:mrms-3csi}
\end{figure}
The first issue with this metric is that different precipitation events inherently respond differently to the thresholds. Figure~\ref{fig:mrms-3csi} depicts three CSI scores of different thresholds for three separate events. As shown, the CSI score response heavily depends on the event's precipitation intensity. While some events respond to all three CSI metric thresholds appropriately (left), others are cut off as precipitation intensity decreases over the lead time (middle), and some do not respond to any CSI metrics (right). The result suggests that for statistical analysis of predictions over the entire test set, we should 1) avoid calculating the average CSI of different thresholds due to the inherently varied responses, and 2) for a specific threshold, only compute the subset of events that respond properly to it.

The second problem is that CSI metric only computes a score for a single time frame in a single event. To extend the CSI metric over the whole test set, we propose \textbf{Cumulative CSI} (CuCSI) which is defined as follows. 
First, CuCSI also requires input parameters, the threshold $T$, the number of bins $N$ and the CSI score step size $s$.
Other than serving the binary categorization purpose, the threshold parameter also selects a subset from the whole test set $\mathbf{X}$, such that: 
\begin{equation}
\mathbf{X_{T}}=\{x \in X \mid \forall_{i=0}^{19}{A_T(x_{i}) + B_T(x_{i}) + C_T(x_{i})} > 0\}
\end{equation}
where $A_T, B_T, C_T$ denote aforementioned items in CSI definition at threshold $T$, and $x_i$ denotes a specific time frame in the event $x$.
We create a 2D grid where each row contains 20 cells representing the lead time, and each column contains $N$ cells of CSI values, starting from 0 and incremented by $s$.
This grid quantizes the CSI scores so that we can compute the cumulated frequencies of CSI scores distributed into the grid cells:
\begin{equation}
\mathrm{CuCSI_T}(i, j)= \lVert \{x_i \in \mathbf{X_{T}} \mid \mathrm{CSI_T}(x_i) \in [sj, s(j+1))\} \rVert
\end{equation}
\begin{figure}\centering
  \includegraphics[width=0.48\textwidth]{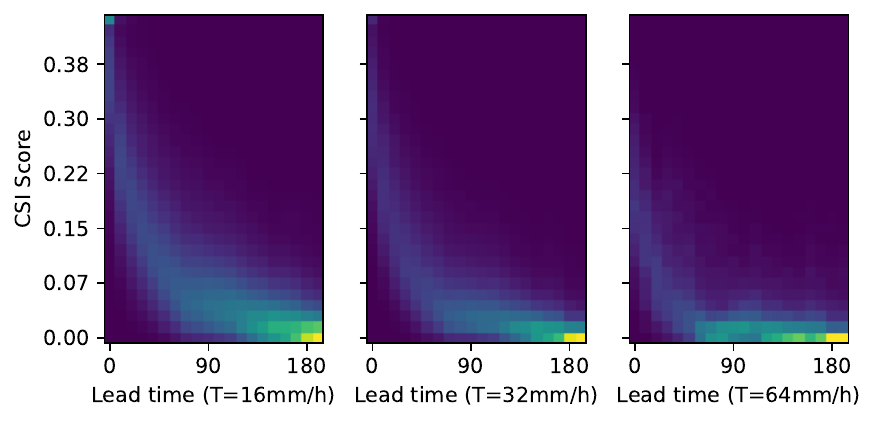}
    \vspace{-20px}
  \caption{Cumulative CSI Over the Whole Dataset}
    \vspace{-10px}
  \label{fig:mrms-cucsi}
\end{figure}
Figure~\ref{fig:mrms-cucsi} illustrates the CuCSI histograms of three different thresholds, $N=30$, $s=0.015$. We can see that the distribution of CuCSI frequencies generally follows an exponential decay trend along X axis (lead time), and higher threshold value leads to faster decay. This implies that more intense precipitation activities are harder to predict. Also, it is worth noting that lower threshold does not imply consistently better performance, but rather a wider spread in the histogram.

\vspace{-8px}
\subsection{Conventional Machine-Learning Metrics}
\begin{figure}\centering
  \includegraphics[width=0.48\textwidth]{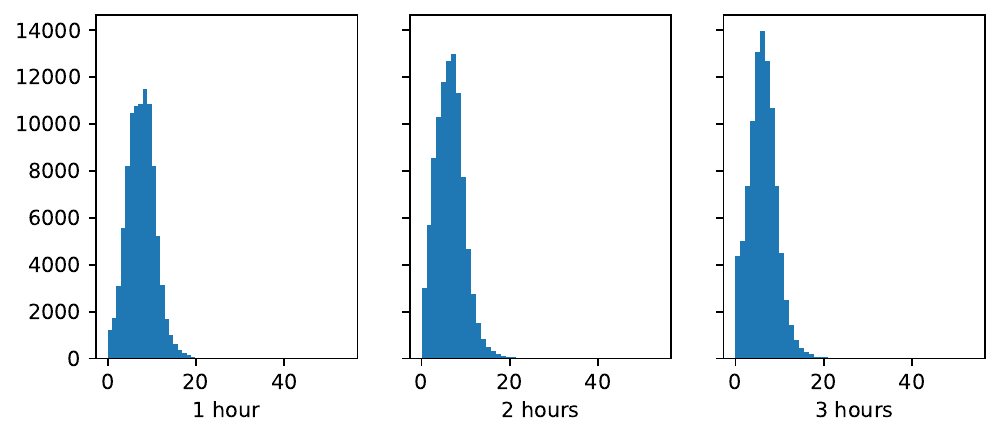}
    \vspace{-20px}
  \caption{Average Precipitation Intensities}
    \vspace{-15px}
  \label{fig:mrms-gt-avg-intensity}
\end{figure}

\begin{table}
  \begin{center}
    \begin{tabular}{l c c c}
    \toprule
       & \textbf{1 Hour} & \textbf{2 hours} & \textbf{3 hours} \\
       \textbf{Raw} & 0.43992 & 0.53529 & 0.52442 \\
       \textbf{Active} & 10.84389 & 11.81609 & 11.96047 \\
    \bottomrule
    \end{tabular}
    \end{center}
    \vspace{-10px}
    \caption{Average MAE of Predictions at Different Lead Time}
    \vspace{-15px}
    \label{table:mrms-avg-mae}
\end{table}
We now examine the model's MAE performance over the entire dataset. Despite the aforementioned issues, this metric is used in the training process as part of the loss function and serves as the ``sink'' of all relevant error source. Therefore, studying this metric and linking it to others is valuable. The MAE metric is typically calculated for each time frame, then averaged across the dataset. Similar to the CuCSI metric, we select time frames from all events at 1, 2, and 3 hours of lead time and compute the average MAE.
A key difference in precipitation prediction compared to other image prediction tasks is the sparsity of the prediction result. Precipitation activities often span a local area, leaving most areas inactive, resulting in many zero readings in the time frames. Unlike conventional image prediction tasks such as ImageNet\cite{dengImageNetLargescaleHierarchical2009}, where zero readings are rare, this can skew the MAE towards a smaller value as significant errors in intensive precipitation predictions are averaged with many small error values. To address this, we propose \textbf{active area MAE}, limiting each time frame's region of interest to an area defined by an intensity threshold. Table\ref{table:mrms-avg-mae} shows both the raw and active area (threshold set to $5\mathrm{mm}/h$) MAE values.
For raw MAE, the model performs consistently across different lead times, with the 3-hour MAE even better than the 2-hour. However, introducing an active area alters this result, showing prediction performance decaying over time. To further understand these seemingly contradictory results, we compute the average precipitation intensity distribution of the events at different time frames, shown in figure~\ref{fig:mrms-gt-avg-intensity}. The histogram provides an evolution overview of the precipitation activities, revealing that many events have decreasing average intensity over time, while some maintain a fairly constant intensity, resulting in a consistent peak at around $10\mathrm{mm}/h$ across all three lead time values. This supports our rationale for introducing the active area MAE metric: events with decreasing average intensity affected the raw MAE, leading to a lower MAE at 3 hours than at 2 hours.

\begin{figure}\centering
  \includegraphics[width=0.48\textwidth]{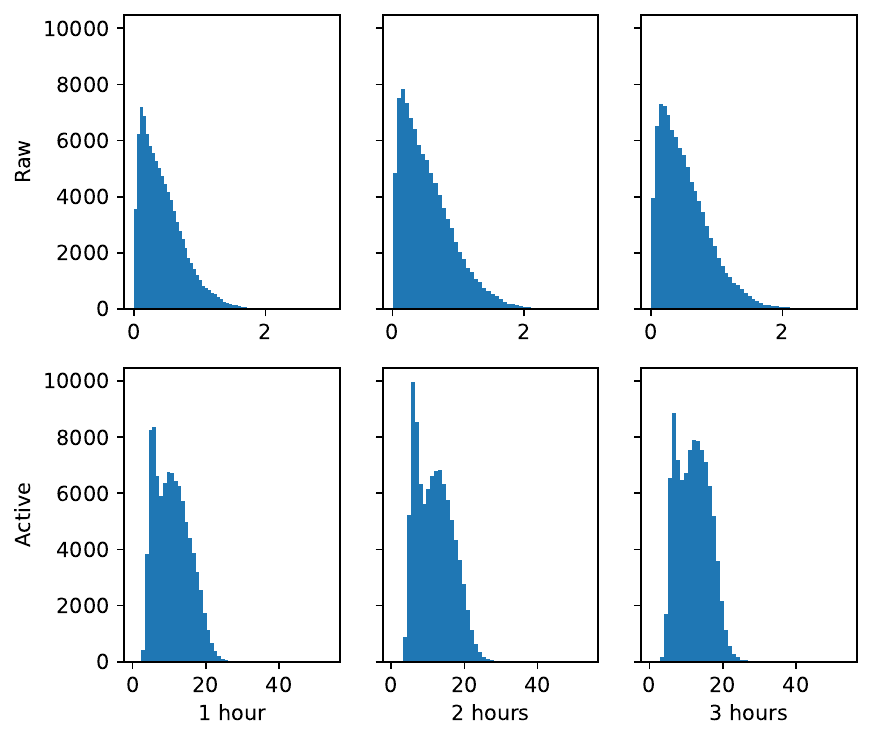}
    \vspace{-20px}
  \caption{The Distribution of MAE And Active Area MAE}
    \vspace{-15px}
  \label{fig:mrms-mae-dist}
\end{figure}

The increased fiedelity of active area MAE metric can be further demonstrated by breaking up the average operator and taking a look at the distribution of MAE values over all events, as shown in figure~\ref{fig:mrms-mae-dist}. The first row of histograms are the raw MAE distribution, second row the active area MAE. We can see that while the raw MAE histograms display a consistent distribution that peaks at low errors, it is caused by the dominance of low intensity local areas. When we filter out such low intensity localities by placing the active area region of interest, a different distribution pattern is revealed, consistently showing two peaks at $6\mathrm{mm}/h$ and $15\mathrm{mm}/h$. The distributions of active area MAE suggests that the test data set contains multiple types of precipitation activities, and the prediction model respond to them differently.
\vspace{-8px}
\subsection{Center of Mass Displacement}
As illustrated in figure~\ref{fig:mrms-moving-mae}, MAE is ineffective for evaluating misaligned prediction results. Here we propose to extract the center of atmospheric mass as a signal, and use the center of mass displacement as a metric to evaluate the alignment performance of the model. Given a precipitation intensity time frame $w$ (a 512x512 grid) and prediction $w'$, the center of mass $(x_c, y_c)$ can be obtained by computing the weighted average coordinates of all pixels, and the displacement metric $\delta_r$ is the distance between the ground truth and prediction centers: 
\vspace{-15px}
\begin{equation}
\begin{split}
x_c(w)=\frac{\sum{j \cdot w_{ij}}}{\sum{w_{ij}}}; y_c(w)=\frac{\sum{i \cdot w_{ij}}}{\sum{w_{ij}}} \\
\delta_r(w, w') = \sqrt{(x_c(w)-x_c(w'))^2 + (y_c(w)-y_c(w'))^2}
\end{split}
\end{equation}
\vspace{-10px}
\begin{figure}\centering
  \includegraphics[width=0.48\textwidth]{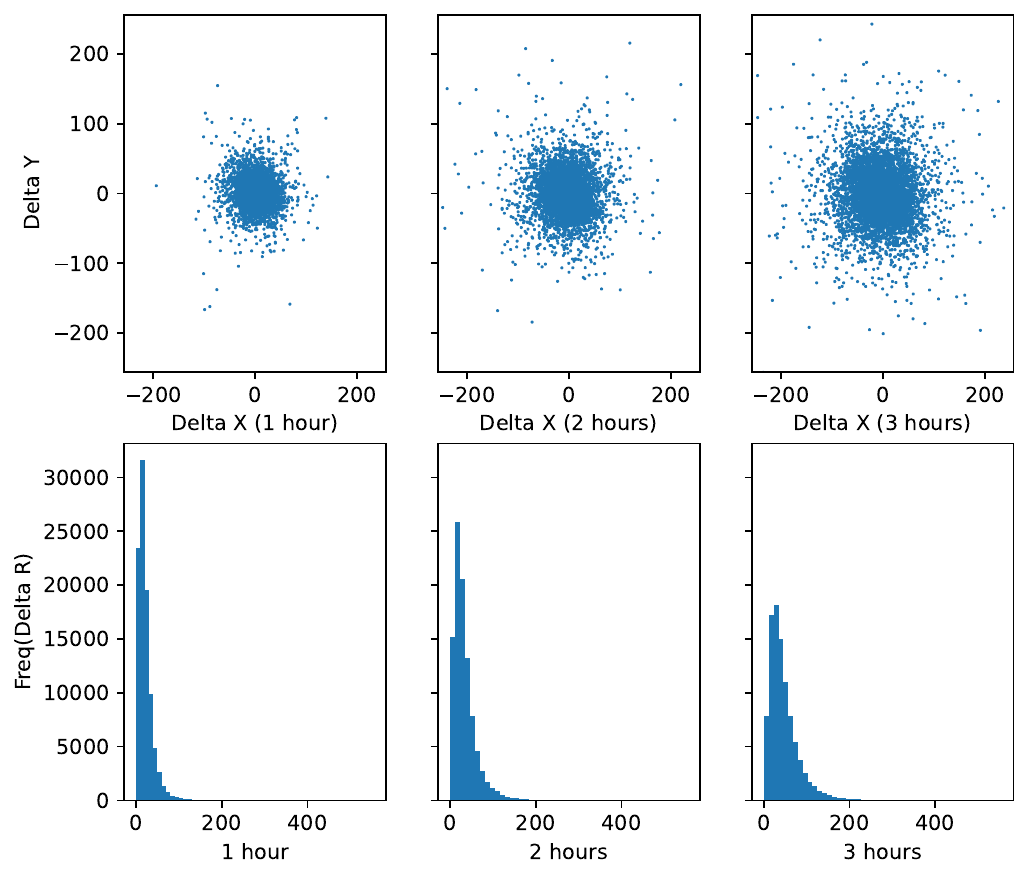}
    \vspace{-20px}
  \caption{Center of Mass Displacement}
    \vspace{-20px}
  \label{fig:mrms-center}
\end{figure}

Figure~\ref{fig:mrms-center} displays the $x_c/y_c$ displacement scatter plots (first row) and $\delta_r$ distribution (second row) for various lead times. The error appears unbiased in X/Y direction, suggesting an unbiased prediction of advection velocity. Both the scatter plot and histogram show that center displacement accumulates over time and is universal in the test set, as evidenced by the histogram peak shifting towards larger values and peak frequencies collapsing into longer tails over time.

\subsection{Differential Trend Analysis}
While the center of mass displacement metric studies the atmospheric mass movement, another crucial aspect of precipitation activity is the total atmospheric mass change, leading to increasing or decreasing precipitation intensity over time. Unlike the aforementioned metrics that compute error on a specific time frame, differential trend analysis focuses on the mass differential introduced by atmospheric activities over a certain period, such as vaporization and precipitation. Given specific start and end time frames denoted by $i$ and $j$, we can obtain the differential trend of ground truth and prediction series as follows:
\vspace{-5px}
\begin{equation}
\mathrm{diff}_{gt}(i, j) = \frac{\sum{x_j} - \sum{x_i}}{512^2}; \mathrm{diff}_{pd}(i, j) = \frac{\sum{x'_j} - \sum{x_i}}{512^2}
\vspace{-10px}
\end{equation}
Note the subtle difference in forms for ground truth and prediction -- we use $x_i$ as a common offset for both at time frame $j$. By offsetting both ground truth and prediction, we eliminate a major correlated component, the ground truth intensity at time $i$, and focus solely on the differential trend, determining whether the precipitation intensifies or weakens. It's also noteworthy that by computing the error of the differential values, the metric degenerates to the average intensity error metric and loses the differential properties, as the offset is cancelled in subtraction.

\begin{figure}\centering
  \includegraphics[width=0.48\textwidth]{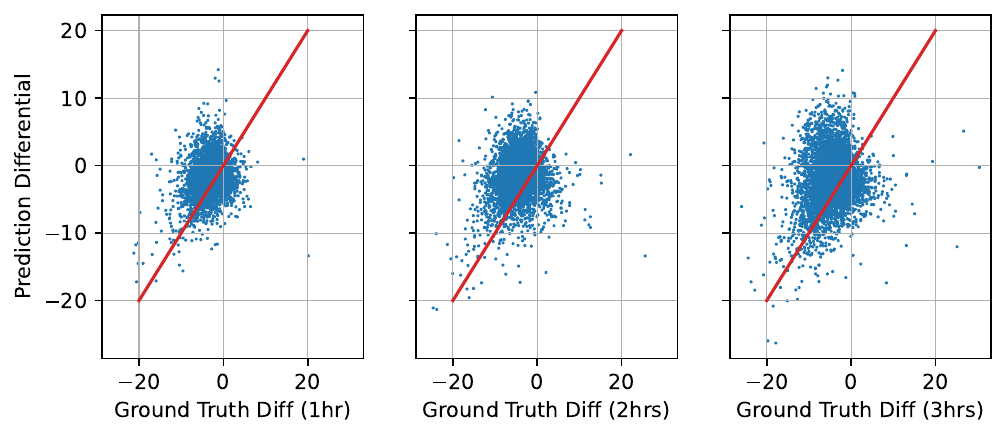}
    \vspace{-20px}
  \caption{Differential Trend Analysis}
    \vspace{-15px}
  \label{fig:mrms-trend}
\end{figure}

Figure~\ref{fig:mrms-trend} presents the scatter plot of $\mathrm{diff}_{gt} / \mathrm{diff}_{pd}$, with a fixed starting time $i=0$, and different end times at 1~3 hours. The X-axis represents the ground truth differential value and the Y-axis the prediction differential. We also plot a red line that represents perfect predictions. The scatter plots reveal an anomaly, with the first quadrant almost empty. Moreover, the plot's right half (quadrants one and four) contains fewer points than the left half. These results imply that the model struggles to accurately predict precipitation events with increasing intensity (first quadrant). It also exposes data imbalance, with more events displaying decreasing intensity over time, resulting in the catastrophic forgetting effect that hinders the model from correctly predicting events with increasing intensity.


\subsection{Prediction Stability Analysis}
NowcastNet significantly differs from other models evaluated in this study because its output is not entirely deterministic. It incorporates a GAN-based decoder to enhance the details. The model architecture first uses a physically constrained evolution module to generate prediction time frames, then an auto-encoder extracts features from the prediction. Finally, the features, along with a Gaussian noise vector, drive the sampling process in the GAN-based decoder's latent space, generating the augmented time frames.

\begin{figure}\centering
  \includegraphics[width=0.48\textwidth]{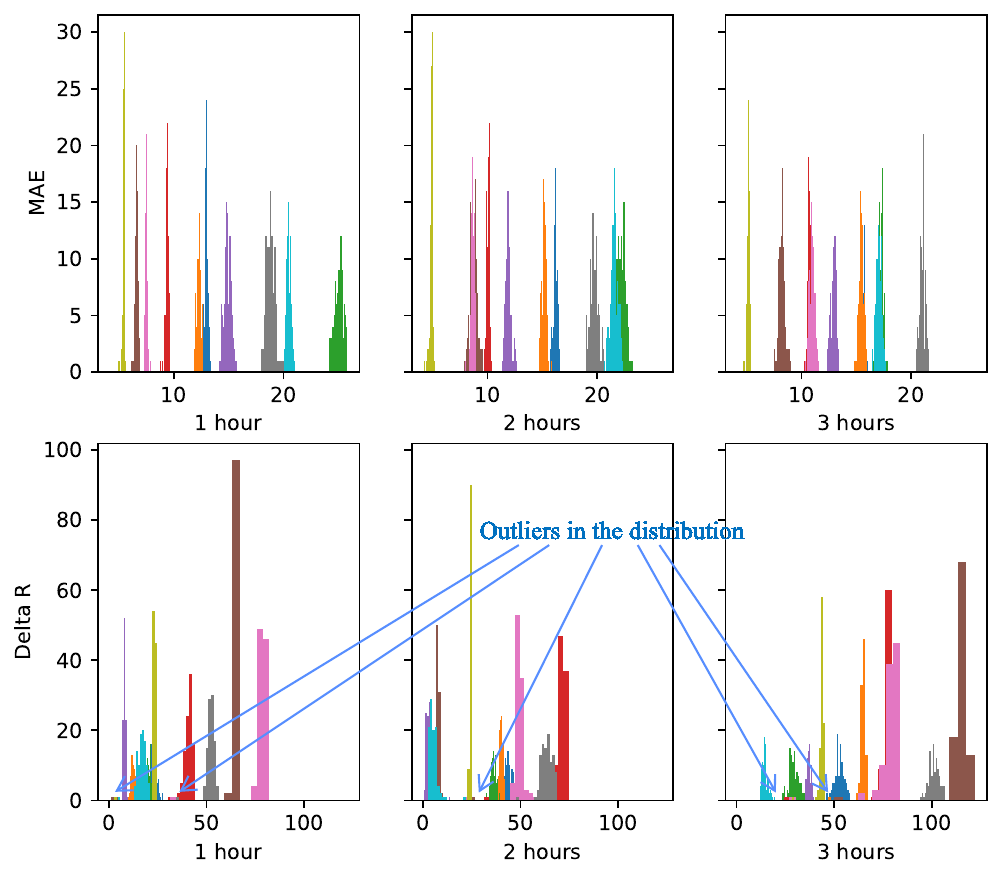}
    \vspace{-25px}
  \caption{Prediction Stability Analysis}
    \vspace{-20px}
  \label{fig:mrms-gan}
\end{figure}

To assess the stability of model predictions, we randomly select 10 events from the test set, run each through the model 100 times. We then compute the MAE and $\delta_r$ metrics, as previously discussed. Figure~\ref{fig:mrms-gan} shows the metrics' distribution for the 10 events over different lead times. Each event is consistently color-coded across all histograms, with 16 bins allocated per event. The bin width for each event varies based on the metric distribution range, wider bins indicating a more spread-out distribution. The figure shows that running the model multiple times on the same event introduces some prediction result uncertainty. The MAE metric distributions are narrow, indicating stable MAE metrics, due to the decoder being conditioned by the evolution module prediction frames with a pixel-to-pixel loss function. However, for the $\delta_r$ histograms, some distributions have wider bins, introduced by significantly deviating prediction results, indicating unstable $\delta_r$ metric across the 100 runs. The $\delta_r$ outliers are marked in blue. The results indicate that while the GAN-based decoder maintains good stability on the total atmospheric mass, it introduces mass distribution instability.

\subsection{Error Tracing}

\begin{figure}\centering
  \includegraphics[width=0.48\textwidth]{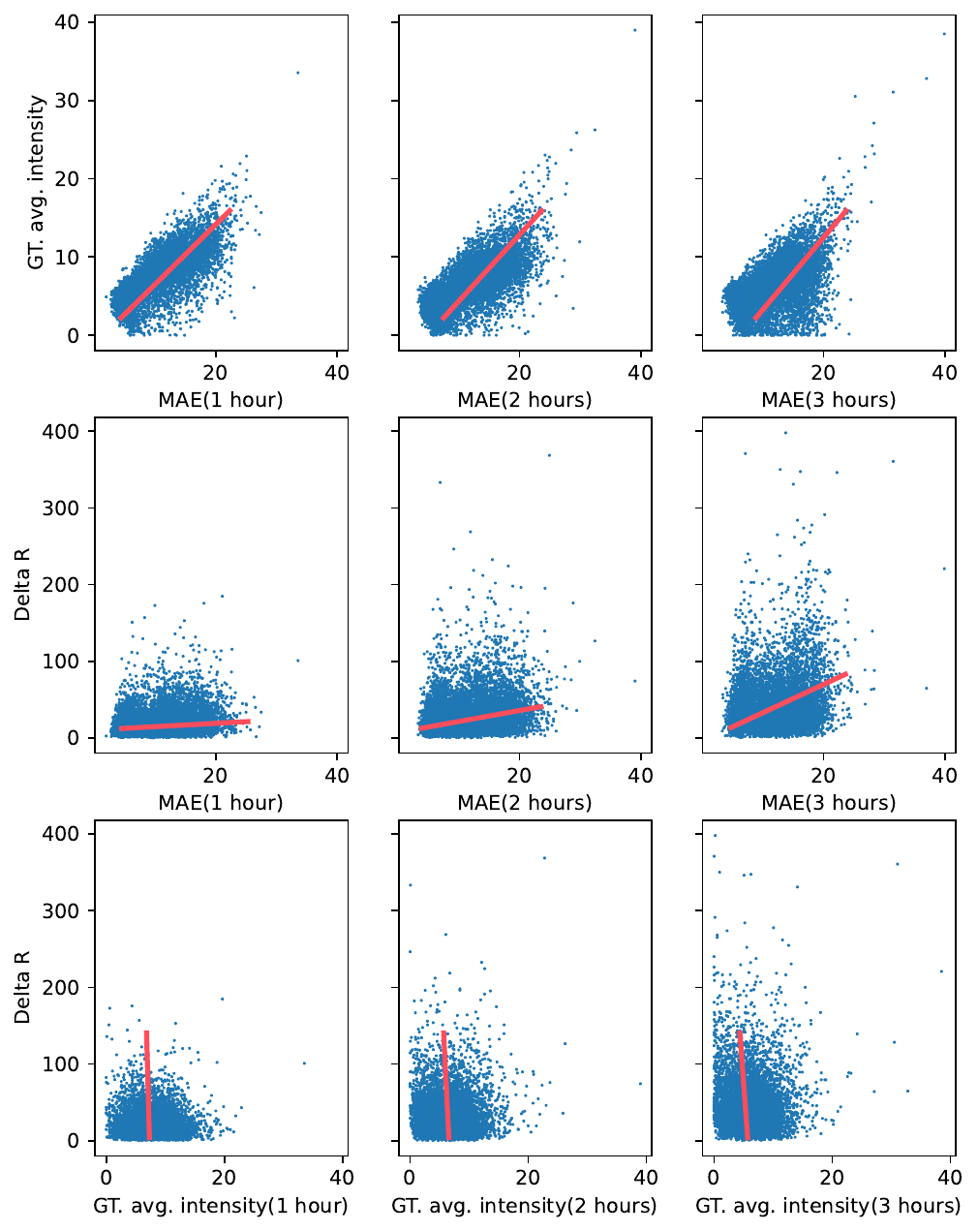}
    \vspace{-15px}
  \caption{Metric Correlation Analysis}
    \vspace{-15px}
  \label{fig:mrms-cor}
\end{figure}

With scientifically meaningful metrics defined, we now explore their relationship and each component's contribution to the error metric that the model optimizes against. To investigate the relationship between MAE, $\delta_r$, and average ground truth intensity, we examine their pairwise correlation, as shown in figure~\ref{fig:mrms-cor}. The data's linear fitting is marked in red for each scatter plot, showing a strong linear relationship between MAE and ground truth average intensity, a weak correlation between MAE and $\delta_r$, and no correlation between $\delta_r$ and ground truth average intensity.
The linear relationship between MAE and ground truth intensity suggests that the intensity error is proportional to the precipitation intensity -- the more extreme the precipitation, the higher the predicted intensity error.
The weak correlation between MAE and $\delta_r$ implies that while the center of mass displacement contributes to the predicted intensity error, there is another unobserved component, whose error also contributes to the MAE metric. Recall the atmospheric mass evolution governing continuity equation (\ref{eq:mrms}), other than the velocity field, the residual component also needs model prediction, so the MAE error can be traced back to two different parts respectively.
The independence between $\delta_r$ and average intensity separates velocity field prediction from ground truth intensity, suggesting that the MAE's proportionality to ground truth intensity results from the proportional error introduced by the unobserved residual prediction.
%


\section{Discussion}

\begin{table}
  \begin{center}
    \begin{tabular}{l l}
    \toprule
       \textbf{Applicable scenario}    & \textbf{Benchmarking technique} \\
       \textbf{Scarce data}            & Sample efficiency \\
       \textbf{Point cloud}            & Rotational invariance \\
                                       & Equivariance test \\
       \textbf{Dense matrix}           & Center-of-mass displacement \\
                                       & Active area scoping \\
                                       & Cumulative metrics \\
       \textbf{Time series}            & Time-domain extrapolation\\
                                       & Time frame slices \\
                                       & Differential trend \\
       \textbf{Scientific features}    & Feature out-of-distribution test \\
       \textbf{Error tracing}          & Per-sample breakdown analysis \\
                                       & Metric correlation analysis \\
       \textbf{Generative model}       & Prediction stability analysis \\
    \bottomrule
    \end{tabular}
    \end{center}
    \vspace{-10px}
    \caption{A Toolbox For Structural Interpretation}
    \label{table:toolbox}
    \vspace{-15px}
\end{table}

As shown in table~\ref{table:workloads}, AI4S is a broad umbrella covering many scientific research fields and workloads. Nevertheless, the workloads share common characteristics that make them more approachable in structural interpretation.
We compile the techniques used in our benchmarking experiments into a ``toolbox'' for structural interpretation, as shown in table~\ref{table:toolbox}. One can mix-and-match suitable techniques for benchmarking a workload, or devise new techniques following the ideas in the toolbox.
Note that the applicability of a technique depends on both the workload scenario and the AI4S model -- for example, we could have done equivariance test on the MLFF workload, but the NequIP model guarantees equivariance against rotation and invariance against translation.

\section{Conclusion}
AI4S is an emerging technique that offers notable enhancements in both computational efficiency and accuracy in comparison to traditional methods. The unique integration of scientific computing and machine learning introduces its own set of challenges. Although the governing natural law plays a pivotal role in data generation, model design, and evaluation, the data-driven nature of the technique can lead to a lack of assurance in terms of correctness and accuracy. Consequently, it is crucial to develop benchmarks that thoroughly analyze the behavior of AI4S models, establish trusted operating ranges, and trace errors. In response to these challenges, we propose an evaluation method known as structural interpretation. This method reveals fine-grained details of model behavior across problem and metric spaces. Benchmarks conducted on three representative AI4S workloads demonstrate that this method can systematically and effectively depict model behavior, both within and beyond available datasets. Furthermore, we provide a comprehensive toolbox of metric design, projection, and correlation discovery techniques that can decouple errors into independent sources and attribute them to corresponding components. Collectively, these techniques provide opportunities to design and develop better models and datasets.

\bibliographystyle{cas-model2-names}

\bibliography{zotero}

\begin{thebibliography}{25}
\expandafter\ifx\csname natexlab\endcsname\relax\def\natexlab#1{#1}\fi
\providecommand{\url}[1]{\texttt{#1}}
\providecommand{\href}[2]{#2}
\providecommand{\path}[1]{#1}
\providecommand{\DOIprefix}{doi:}
\providecommand{\ArXivprefix}{arXiv:}
\providecommand{\URLprefix}{URL: }
\providecommand{\Pubmedprefix}{pmid:}
\providecommand{\doi}[1]{\href{http://dx.doi.org/#1}{\path{#1}}}
\providecommand{\Pubmed}[1]{\href{pmid:#1}{\path{#1}}}
\providecommand{\bibinfo}[2]{#2}
\ifx\xfnm\relax \def\xfnm[#1]{\unskip,\space#1}\fi
\bibitem[{Bart{\'o}k et~al.(2013)Bart{\'o}k, Kondor and
  Cs{\'a}nyi}]{bartokRepresentingChemicalEnvironments2013}
\bibinfo{author}{Bart{\'o}k, A.P.}, \bibinfo{author}{Kondor, R.},
  \bibinfo{author}{Cs{\'a}nyi, G.}, \bibinfo{year}{2013}.
\newblock \bibinfo{title}{On representing chemical environments}.
\newblock \bibinfo{journal}{Physical Review B} \bibinfo{volume}{87},
  \bibinfo{pages}{184115}.
\newblock \URLprefix \url{https://link.aps.org/doi/10.1103/PhysRevB.87.184115},
  \DOIprefix\doi{10.1103/PhysRevB.87.184115}.
\bibitem[{Bart{\'o}k et~al.(2010)Bart{\'o}k, Payne, Kondor and
  Cs{\'a}nyi}]{bartokGaussianApproximationPotentials2010}
\bibinfo{author}{Bart{\'o}k, A.P.}, \bibinfo{author}{Payne, M.C.},
  \bibinfo{author}{Kondor, R.}, \bibinfo{author}{Cs{\'a}nyi, G.},
  \bibinfo{year}{2010}.
\newblock \bibinfo{title}{Gaussian {{Approximation Potentials}}: {{The
  Accuracy}} of {{Quantum Mechanics}}, without the {{Electrons}}}.
\newblock \bibinfo{journal}{Physical Review Letters} \bibinfo{volume}{104},
  \bibinfo{pages}{136403}.
\newblock \URLprefix
  \url{https://link.aps.org/doi/10.1103/PhysRevLett.104.136403},
  \DOIprefix\doi{10.1103/PhysRevLett.104.136403}.
\bibitem[{Batzner et~al.(2022)Batzner, Musaelian, Sun, Geiger, Mailoa,
  Kornbluth, Molinari, Smidt and Kozinsky}]{batznerEquivariantGraphNeural2022}
\bibinfo{author}{Batzner, S.}, \bibinfo{author}{Musaelian, A.},
  \bibinfo{author}{Sun, L.}, \bibinfo{author}{Geiger, M.},
  \bibinfo{author}{Mailoa, J.P.}, \bibinfo{author}{Kornbluth, M.},
  \bibinfo{author}{Molinari, N.}, \bibinfo{author}{Smidt, T.E.},
  \bibinfo{author}{Kozinsky, B.}, \bibinfo{year}{2022}.
\newblock \bibinfo{title}{E(3)-equivariant graph neural networks for
  data-efficient and accurate interatomic potentials}.
\newblock \bibinfo{journal}{Nature Communications} \bibinfo{volume}{13},
  \bibinfo{pages}{2453}.
\newblock \URLprefix \url{https://www.nature.com/articles/s41467-022-29939-5},
  \DOIprefix\doi{10.1038/s41467-022-29939-5}.
\bibitem[{Behler and
  Parrinello(2007)}]{behlerGeneralizedNeuralNetworkRepresentation2007}
\bibinfo{author}{Behler, J.}, \bibinfo{author}{Parrinello, M.},
  \bibinfo{year}{2007}.
\newblock \bibinfo{title}{Generalized {{Neural-Network Representation}} of
  {{High-Dimensional Potential-Energy Surfaces}}}.
\newblock \bibinfo{journal}{Physical Review Letters} \bibinfo{volume}{98},
  \bibinfo{pages}{146401}.
\newblock \URLprefix
  \url{https://link.aps.org/doi/10.1103/PhysRevLett.98.146401},
  \DOIprefix\doi{10.1103/PhysRevLett.98.146401}.
\bibitem[{Chmiela et~al.(2017)Chmiela, Tkatchenko, Sauceda, Poltavsky,
  Sch{\"u}tt and M{\"u}ller}]{chmielaMachineLearningAccurate2017}
\bibinfo{author}{Chmiela, S.}, \bibinfo{author}{Tkatchenko, A.},
  \bibinfo{author}{Sauceda, H.E.}, \bibinfo{author}{Poltavsky, I.},
  \bibinfo{author}{Sch{\"u}tt, K.T.}, \bibinfo{author}{M{\"u}ller, K.R.},
  \bibinfo{year}{2017}.
\newblock \bibinfo{title}{Machine learning of accurate energy-conserving
  molecular force fields}.
\newblock \bibinfo{journal}{Science Advances} \bibinfo{volume}{3},
  \bibinfo{pages}{e1603015}.
\newblock \URLprefix \url{https://www.science.org/doi/10.1126/sciadv.1603015},
  \DOIprefix\doi{10.1126/sciadv.1603015}.
\bibitem[{Christensen and {von
  Lilienfeld}(2020)}]{christensenRoleGradientsMachine2020}
\bibinfo{author}{Christensen, A.S.}, \bibinfo{author}{{von Lilienfeld}, O.A.},
  \bibinfo{year}{2020}.
\newblock \bibinfo{title}{On the role of gradients for machine learning of
  molecular energies and forces}.
\newblock \URLprefix \url{http://arxiv.org/abs/2007.09593},
  \href{http://arxiv.org/abs/2007.09593}{\tt arXiv:2007.09593}.
\bibitem[{Deng et~al.(2009)Deng, Dong, Socher, Li, Li and
  {Fei-Fei}}]{dengImageNetLargescaleHierarchical2009}
\bibinfo{author}{Deng, J.}, \bibinfo{author}{Dong, W.},
  \bibinfo{author}{Socher, R.}, \bibinfo{author}{Li, L.J.},
  \bibinfo{author}{Li, K.}, \bibinfo{author}{{Fei-Fei}, L.},
  \bibinfo{year}{2009}.
\newblock \bibinfo{title}{{{ImageNet}}: {{A}} large-scale hierarchical image
  database}, in: \bibinfo{booktitle}{2009 {{IEEE Conference}} on {{Computer
  Vision}} and {{Pattern Recognition}}}, pp. \bibinfo{pages}{248--255}.
\newblock \DOIprefix\doi{10.1109/CVPR.2009.5206848}.
\bibitem[{Gao et~al.(2019a)Gao, Luo, Wang, Xiong, Chen, Hao, Jiang, Fan, Du,
  Huang, Zhang, Wen, Zheng, He, Dai, Ye, Cao, Jia, Zhan, Tang, Zheng, Xie, Li,
  Wang and Zhan}]{gaoAIBenchScalableComprehensive2019}
\bibinfo{author}{Gao, W.}, \bibinfo{author}{Luo, C.}, \bibinfo{author}{Wang,
  L.}, \bibinfo{author}{Xiong, X.}, \bibinfo{author}{Chen, J.},
  \bibinfo{author}{Hao, T.}, \bibinfo{author}{Jiang, Z.}, \bibinfo{author}{Fan,
  F.}, \bibinfo{author}{Du, M.}, \bibinfo{author}{Huang, Y.},
  \bibinfo{author}{Zhang, F.}, \bibinfo{author}{Wen, X.},
  \bibinfo{author}{Zheng, C.}, \bibinfo{author}{He, X.}, \bibinfo{author}{Dai,
  J.}, \bibinfo{author}{Ye, H.}, \bibinfo{author}{Cao, Z.},
  \bibinfo{author}{Jia, Z.}, \bibinfo{author}{Zhan, K.}, \bibinfo{author}{Tang,
  H.}, \bibinfo{author}{Zheng, D.}, \bibinfo{author}{Xie, B.},
  \bibinfo{author}{Li, W.}, \bibinfo{author}{Wang, X.}, \bibinfo{author}{Zhan,
  J.}, \bibinfo{year}{2019}a.
\newblock \bibinfo{title}{{{AIBench}}: {{Towards Scalable}} and {{Comprehensive
  Datacenter AI Benchmarking}}}, in: \bibinfo{editor}{Zheng, C.},
  \bibinfo{editor}{Zhan, J.} (Eds.), \bibinfo{booktitle}{Benchmarking,
  {{Measuring}}, and {{Optimizing}}}. \bibinfo{publisher}{{Springer
  International Publishing}}, \bibinfo{address}{{Cham}}. volume
  \bibinfo{volume}{11459}, pp. \bibinfo{pages}{3--9}.
\newblock \URLprefix
  \url{http://link.springer.com/10.1007/978-3-030-32813-9_1},
  \DOIprefix\doi{10.1007/978-3-030-32813-9_1}.
\bibitem[{Gao et~al.(2019b)Gao, Tang, Wang, Zhan, Lan, Luo, Huang, Zheng, Dai,
  Cao, Zheng, Tang, Zhan, Wang, Kong, Wu, Yu, Tan, Li, Tian, Li, Shao, Wang,
  Wang and Ye}]{gaoAIBenchIndustryStandard2019}
\bibinfo{author}{Gao, W.}, \bibinfo{author}{Tang, F.}, \bibinfo{author}{Wang,
  L.}, \bibinfo{author}{Zhan, J.}, \bibinfo{author}{Lan, C.},
  \bibinfo{author}{Luo, C.}, \bibinfo{author}{Huang, Y.},
  \bibinfo{author}{Zheng, C.}, \bibinfo{author}{Dai, J.}, \bibinfo{author}{Cao,
  Z.}, \bibinfo{author}{Zheng, D.}, \bibinfo{author}{Tang, H.},
  \bibinfo{author}{Zhan, K.}, \bibinfo{author}{Wang, B.},
  \bibinfo{author}{Kong, D.}, \bibinfo{author}{Wu, T.}, \bibinfo{author}{Yu,
  M.}, \bibinfo{author}{Tan, C.}, \bibinfo{author}{Li, H.},
  \bibinfo{author}{Tian, X.}, \bibinfo{author}{Li, Y.}, \bibinfo{author}{Shao,
  J.}, \bibinfo{author}{Wang, Z.}, \bibinfo{author}{Wang, X.},
  \bibinfo{author}{Ye, H.}, \bibinfo{year}{2019}b.
\newblock \bibinfo{title}{{{AIBench}}: {{An Industry Standard Internet Service
  AI Benchmark Suite}}}.
\newblock \bibinfo{journal}{arXiv:1908.08998 [cs]} \URLprefix
  \url{http://arxiv.org/abs/1908.08998},
  \href{http://arxiv.org/abs/1908.08998}{\tt arXiv:1908.08998}.
\bibitem[{Gao et~al.(2021)Gao, Tang, Zhan, Wen, Wang, Cao, Lan, Luo, Liu and
  Jiang}]{gaoAIBenchScenarioScenarioDistilling2021}
\bibinfo{author}{Gao, W.}, \bibinfo{author}{Tang, F.}, \bibinfo{author}{Zhan,
  J.}, \bibinfo{author}{Wen, X.}, \bibinfo{author}{Wang, L.},
  \bibinfo{author}{Cao, Z.}, \bibinfo{author}{Lan, C.}, \bibinfo{author}{Luo,
  C.}, \bibinfo{author}{Liu, X.}, \bibinfo{author}{Jiang, Z.},
  \bibinfo{year}{2021}.
\newblock \bibinfo{title}{{{AIBench Scenario}}: {{Scenario-Distilling AI
  Benchmarking}}}, in: \bibinfo{booktitle}{2021 30th {{International
  Conference}} on {{Parallel Architectures}} and {{Compilation Techniques}}
  ({{PACT}})}, \bibinfo{publisher}{{IEEE}}, \bibinfo{address}{{Atlanta, GA,
  USA}}. pp. \bibinfo{pages}{142--158}.
\newblock \URLprefix \url{https://ieeexplore.ieee.org/document/9563026/},
  \DOIprefix\doi{10.1109/PACT52795.2021.00018}.
\bibitem[{Gohel et~al.(2021)Gohel, Singh and
  Mohanty}]{gohelExplainableAICurrent2021}
\bibinfo{author}{Gohel, P.}, \bibinfo{author}{Singh, P.},
  \bibinfo{author}{Mohanty, M.}, \bibinfo{year}{2021}.
\newblock \bibinfo{title}{Explainable {{AI}}: Current status and future
  directions}.
\newblock \URLprefix \url{http://arxiv.org/abs/2107.07045},
  \DOIprefix\doi{10.48550/arXiv.2107.07045},
  \href{http://arxiv.org/abs/2107.07045}{\tt arXiv:2107.07045}.
\bibitem[{Hogan and Mason(2011)}]{hoganDeterministicForecastsBinary2011}
\bibinfo{author}{Hogan, R.J.}, \bibinfo{author}{Mason, I.B.},
  \bibinfo{year}{2011}.
\newblock \bibinfo{title}{Deterministic {{Forecasts}} of {{Binary Events}}},
  in: \bibinfo{booktitle}{Forecast {{Verification}}}. \bibinfo{publisher}{{John
  Wiley \& Sons, Ltd}}. chapter~\bibinfo{chapter}{3}, pp.
  \bibinfo{pages}{31--59}.
\newblock \URLprefix
  \url{https://onlinelibrary.wiley.com/doi/abs/10.1002/9781119960003.ch3},
  \DOIprefix\doi{10.1002/9781119960003.ch3}.
\bibitem[{Kasieczka et~al.(2019)Kasieczka, Plehn, Butter, Cranmer, Debnath,
  Dillon, Fairbairn, Faroughy, Fedorko, Gay, Gouskos, F.~Kamenik, Komiske,
  Leiss, Lister, Macaluso, Metodiev, Moore, Nachman, Nordstr{\"o}m, Pearkes,
  Qu, Rath, Rieger, Shih, Thompson and
  Varma}]{kasieczkaMachineLearningLandscape2019}
\bibinfo{author}{Kasieczka, G.}, \bibinfo{author}{Plehn, T.},
  \bibinfo{author}{Butter, A.}, \bibinfo{author}{Cranmer, K.},
  \bibinfo{author}{Debnath, D.}, \bibinfo{author}{Dillon, B.M.},
  \bibinfo{author}{Fairbairn, M.}, \bibinfo{author}{Faroughy, D.A.},
  \bibinfo{author}{Fedorko, W.}, \bibinfo{author}{Gay, C.},
  \bibinfo{author}{Gouskos, L.}, \bibinfo{author}{F.~Kamenik, J.},
  \bibinfo{author}{Komiske, P.}, \bibinfo{author}{Leiss, S.},
  \bibinfo{author}{Lister, A.}, \bibinfo{author}{Macaluso, S.},
  \bibinfo{author}{Metodiev, E.}, \bibinfo{author}{Moore, L.},
  \bibinfo{author}{Nachman, B.}, \bibinfo{author}{Nordstr{\"o}m, K.},
  \bibinfo{author}{Pearkes, J.}, \bibinfo{author}{Qu, H.},
  \bibinfo{author}{Rath, Y.}, \bibinfo{author}{Rieger, M.},
  \bibinfo{author}{Shih, D.}, \bibinfo{author}{Thompson, J.},
  \bibinfo{author}{Varma, S.}, \bibinfo{year}{2019}.
\newblock \bibinfo{title}{The {{Machine Learning}} landscape of top taggers}.
\newblock \bibinfo{journal}{SciPost Physics} \bibinfo{volume}{7},
  \bibinfo{pages}{014}.
\newblock \URLprefix \url{https://scipost.org/10.21468/SciPostPhys.7.1.014},
  \DOIprefix\doi{10.21468/SciPostPhys.7.1.014}.
\bibitem[{Laboratory()}]{argonnenationallaboratoryAIScienceReport}
\bibinfo{author}{Laboratory, A.N.}, .
\newblock \bibinfo{title}{{{AI}} for {{Science Report}}}.
\newblock \bibinfo{type}{Technical Report}.
\newblock \URLprefix
  \url{https://publications.anl.gov/anlpubs/2020/03/158802.pdf}.
\bibitem[{Li et~al.(2023)Li, Gao, Wang, Sun, Wang and
  Zhan}]{liDoesAIScience2023}
\bibinfo{author}{Li, Y.}, \bibinfo{author}{Gao, W.}, \bibinfo{author}{Wang,
  L.}, \bibinfo{author}{Sun, L.}, \bibinfo{author}{Wang, Z.},
  \bibinfo{author}{Zhan, J.}, \bibinfo{year}{2023}.
\newblock \bibinfo{title}{Does {{AI}} for science need another {{ImageNet Or}}
  totally different benchmarks? {{A}} case study of machine learning force
  fields}.
\newblock \URLprefix \url{http://arxiv.org/abs/2308.05999},
  \DOIprefix\doi{10.48550/arXiv.2308.05999},
  \href{http://arxiv.org/abs/2308.05999}{\tt arXiv:2308.05999}.
\bibitem[{Mattson et~al.()Mattson, Cheng, Coleman, Diamos, Micikevicius,
  Patterson, Tang, Wei, Bailis, Bittorf, Brooks, Chen, Dutta, Gupta, Hazelwood,
  Hock, Huang, Ike, Jia, Kang, Kanter, Kumar, Liao, Ma, Narayanan, Oguntebi,
  Pekhimenko, Pentecost, Reddi, Robie, John, Tabaru, Wu, Xu, Yamazaki, Young
  and Zaharia}]{mattsonMLPerfTrainingBenchmark}
\bibinfo{author}{Mattson, P.}, \bibinfo{author}{Cheng, C.},
  \bibinfo{author}{Coleman, C.}, \bibinfo{author}{Diamos, G.},
  \bibinfo{author}{Micikevicius, P.}, \bibinfo{author}{Patterson, D.},
  \bibinfo{author}{Tang, H.}, \bibinfo{author}{Wei, G.Y.},
  \bibinfo{author}{Bailis, P.}, \bibinfo{author}{Bittorf, V.},
  \bibinfo{author}{Brooks, D.}, \bibinfo{author}{Chen, D.},
  \bibinfo{author}{Dutta, D.}, \bibinfo{author}{Gupta, U.},
  \bibinfo{author}{Hazelwood, K.}, \bibinfo{author}{Hock, A.},
  \bibinfo{author}{Huang, X.}, \bibinfo{author}{Ike, A.}, \bibinfo{author}{Jia,
  B.}, \bibinfo{author}{Kang, D.}, \bibinfo{author}{Kanter, D.},
  \bibinfo{author}{Kumar, N.}, \bibinfo{author}{Liao, J.}, \bibinfo{author}{Ma,
  G.}, \bibinfo{author}{Narayanan, D.}, \bibinfo{author}{Oguntebi, T.},
  \bibinfo{author}{Pekhimenko, G.}, \bibinfo{author}{Pentecost, L.},
  \bibinfo{author}{Reddi, V.J.}, \bibinfo{author}{Robie, T.},
  \bibinfo{author}{John, T.S.}, \bibinfo{author}{Tabaru, T.},
  \bibinfo{author}{Wu, C.J.}, \bibinfo{author}{Xu, L.},
  \bibinfo{author}{Yamazaki, M.}, \bibinfo{author}{Young, C.},
  \bibinfo{author}{Zaharia, M.}, .
\newblock \bibinfo{title}{{{MLPerf Training Benchmark}}} , \bibinfo{pages}{14}.
\bibitem[{Qu and Gouskos(2020)}]{quParticlenetJetTagging2020}
\bibinfo{author}{Qu, H.}, \bibinfo{author}{Gouskos, L.}, \bibinfo{year}{2020}.
\newblock \bibinfo{title}{[particlenet] {{Jet}} tagging via particle clouds}.
\newblock \bibinfo{journal}{Physical Review D} \bibinfo{volume}{101},
  \bibinfo{pages}{056019}.
\newblock \URLprefix
  \url{https://link.aps.org/doi/10.1103/PhysRevD.101.056019},
  \DOIprefix\doi{10.1103/PhysRevD.101.056019}.
\bibitem[{Qu et~al.(2022)Qu, Li and Qian}]{quParticleTransformerJet2022}
\bibinfo{author}{Qu, H.}, \bibinfo{author}{Li, C.}, \bibinfo{author}{Qian, S.},
  \bibinfo{year}{2022}.
\newblock \bibinfo{title}{Particle {{Transformer}} for {{Jet Tagging}}}.
\newblock \URLprefix \url{http://arxiv.org/abs/2202.03772},
  \DOIprefix\doi{10.48550/arXiv.2202.03772},
  \href{http://arxiv.org/abs/2202.03772}{\tt arXiv:2202.03772}.
\bibitem[{Racah et~al.(2017)Racah, Beckham, Maharaj, Ebrahimi~Kahou, Prabhat
  and Pal}]{racahExtremeWeatherLargescaleClimate2017}
\bibinfo{author}{Racah, E.}, \bibinfo{author}{Beckham, C.},
  \bibinfo{author}{Maharaj, T.}, \bibinfo{author}{Ebrahimi~Kahou, S.},
  \bibinfo{author}{Prabhat, {\relax Mr}.}, \bibinfo{author}{Pal, C.},
  \bibinfo{year}{2017}.
\newblock \bibinfo{title}{{{ExtremeWeather}}: {{A}} large-scale climate dataset
  for semi-supervised detection, localization, and understanding of extreme
  weather events}, in: \bibinfo{booktitle}{Advances in {{Neural Information
  Processing Systems}}}, \bibinfo{publisher}{{Curran Associates, Inc.}}
\newblock \URLprefix
  \url{https://proceedings.neurips.cc/paper/2017/hash/519c84155964659375821f7ca576f095-Abstract.html}.
\bibitem[{Ravuri et~al.(2021)Ravuri, Lenc, Willson, Kangin, Lam, Mirowski,
  Fitzsimons, Athanassiadou, Kashem, Madge, Prudden, Mandhane, Clark, Brock,
  Simonyan, Hadsell, Robinson, Clancy, Arribas and
  Mohamed}]{ravuriSkilfulPrecipitationNowcasting2021}
\bibinfo{author}{Ravuri, S.}, \bibinfo{author}{Lenc, K.},
  \bibinfo{author}{Willson, M.}, \bibinfo{author}{Kangin, D.},
  \bibinfo{author}{Lam, R.}, \bibinfo{author}{Mirowski, P.},
  \bibinfo{author}{Fitzsimons, M.}, \bibinfo{author}{Athanassiadou, M.},
  \bibinfo{author}{Kashem, S.}, \bibinfo{author}{Madge, S.},
  \bibinfo{author}{Prudden, R.}, \bibinfo{author}{Mandhane, A.},
  \bibinfo{author}{Clark, A.}, \bibinfo{author}{Brock, A.},
  \bibinfo{author}{Simonyan, K.}, \bibinfo{author}{Hadsell, R.},
  \bibinfo{author}{Robinson, N.}, \bibinfo{author}{Clancy, E.},
  \bibinfo{author}{Arribas, A.}, \bibinfo{author}{Mohamed, S.},
  \bibinfo{year}{2021}.
\newblock \bibinfo{title}{Skilful precipitation nowcasting using deep
  generative models of radar}.
\newblock \bibinfo{journal}{Nature} \bibinfo{volume}{597},
  \bibinfo{pages}{672--677}.
\newblock \URLprefix \url{https://www.nature.com/articles/s41586-021-03854-z},
  \DOIprefix\doi{10.1038/s41586-021-03854-z}.
\bibitem[{Thiyagalingam et~al.(2022)Thiyagalingam, Shankar, Fox and
  Hey}]{thiyagalingamScientificMachineLearning2022}
\bibinfo{author}{Thiyagalingam, J.}, \bibinfo{author}{Shankar, M.},
  \bibinfo{author}{Fox, G.}, \bibinfo{author}{Hey, T.}, \bibinfo{year}{2022}.
\newblock \bibinfo{title}{Scientific machine learning benchmarks}.
\newblock \bibinfo{journal}{Nature Reviews Physics} \bibinfo{volume}{4},
  \bibinfo{pages}{413--420}.
\newblock \URLprefix \url{https://www.nature.com/articles/s42254-022-00441-7},
  \DOIprefix\doi{10.1038/s42254-022-00441-7}.
\bibitem[{Unke and Meuwly(2019)}]{unkePhysNetNeuralNetwork2019}
\bibinfo{author}{Unke, O.T.}, \bibinfo{author}{Meuwly, M.},
  \bibinfo{year}{2019}.
\newblock \bibinfo{title}{{{PhysNet}}: {{A Neural Network}} for {{Predicting
  Energies}}, {{Forces}}, {{Dipole Moments}} and {{Partial Charges}}}.
\newblock \bibinfo{journal}{Journal of Chemical Theory and Computation}
  \bibinfo{volume}{15}, \bibinfo{pages}{3678--3693}.
\newblock \URLprefix \url{http://arxiv.org/abs/1902.08408},
  \DOIprefix\doi{10.1021/acs.jctc.9b00181},
  \href{http://arxiv.org/abs/1902.08408}{\tt arXiv:1902.08408}.
\bibitem[{Wang et~al.(2017)Wang, Long, Wang, Gao and
  Yu}]{wangPredRNNRecurrentNeural2017}
\bibinfo{author}{Wang, Y.}, \bibinfo{author}{Long, M.}, \bibinfo{author}{Wang,
  J.}, \bibinfo{author}{Gao, Z.}, \bibinfo{author}{Yu, P.S.},
  \bibinfo{year}{2017}.
\newblock \bibinfo{title}{{{PredRNN}}: {{Recurrent Neural Networks}} for
  {{Predictive Learning}} using {{Spatiotemporal LSTMs}}}, in:
  \bibinfo{booktitle}{Advances in {{Neural Information Processing Systems}}},
  \bibinfo{publisher}{{Curran Associates, Inc.}}
\newblock \URLprefix
  \url{https://papers.nips.cc/paper_files/paper/2017/hash/e5f6ad6ce374177eef023bf5d0c018b6-Abstract.html}.
\bibitem[{Zhang et~al.(2016)Zhang, Howard, Langston, Kaney, Qi, Tang, Grams,
  Wang, Cocks, Martinaitis, Arthur, Cooper, Brogden and
  Kitzmiller}]{zhangMultiRadarMultiSensorMRMS2016}
\bibinfo{author}{Zhang, J.}, \bibinfo{author}{Howard, K.},
  \bibinfo{author}{Langston, C.}, \bibinfo{author}{Kaney, B.},
  \bibinfo{author}{Qi, Y.}, \bibinfo{author}{Tang, L.}, \bibinfo{author}{Grams,
  H.}, \bibinfo{author}{Wang, Y.}, \bibinfo{author}{Cocks, S.},
  \bibinfo{author}{Martinaitis, S.}, \bibinfo{author}{Arthur, A.},
  \bibinfo{author}{Cooper, K.}, \bibinfo{author}{Brogden, J.},
  \bibinfo{author}{Kitzmiller, D.}, \bibinfo{year}{2016}.
\newblock \bibinfo{title}{Multi-{{Radar Multi-Sensor}} ({{MRMS}})
  {{Quantitative Precipitation Estimation}}: {{Initial Operating
  Capabilities}}}.
\newblock \bibinfo{journal}{Bulletin of the American Meteorological Society}
  \bibinfo{volume}{97}, \bibinfo{pages}{621--638}.
\newblock \URLprefix
  \url{https://journals.ametsoc.org/view/journals/bams/97/4/bams-d-14-00174.1.xml},
  \DOIprefix\doi{10.1175/BAMS-D-14-00174.1}.
\bibitem[{Zhang et~al.(2023)Zhang, Long, Chen, Xing, Jin, Jordan and
  Wang}]{zhangSkilfulNowcastingExtreme2023}
\bibinfo{author}{Zhang, Y.}, \bibinfo{author}{Long, M.}, \bibinfo{author}{Chen,
  K.}, \bibinfo{author}{Xing, L.}, \bibinfo{author}{Jin, R.},
  \bibinfo{author}{Jordan, M.I.}, \bibinfo{author}{Wang, J.},
  \bibinfo{year}{2023}.
\newblock \bibinfo{title}{Skilful nowcasting of extreme precipitation with
  {{NowcastNet}}}.
\newblock \bibinfo{journal}{Nature} \bibinfo{volume}{619},
  \bibinfo{pages}{526--532}.
\newblock \URLprefix \url{https://www.nature.com/articles/s41586-023-06184-4},
  \DOIprefix\doi{10.1038/s41586-023-06184-4}.

\end{thebibliography}

\end{document}